\PassOptionsToPackage{table}{xcolor}

\documentclass{article} 
\usepackage{iclr2026_conference,times}

\usepackage{amsmath,amsfonts,bm}

\def\eqref#1{equation~\ref{#1}}

\def\1{\bm{1}}

\def\ra{{\textnormal{a}}}

\def\rx{{\textnormal{x}}}

\def\rva{{\mathbf{a}}}

\def\erva{{\textnormal{a}}}

\def\ervx{{\textnormal{x}}}

\def\rmA{{\mathbf{A}}}

\def\vmu{{\bm{\mu}}}
\def\vtheta{{\bm{\theta}}}
\def\va{{\bm{a}}}

\def\ve{{\bm{e}}}

\def\vx{{\bm{x}}}

\def\eva{{a}}

\def\mA{{\bm{A}}}

\def\mH{{\bm{H}}}
\def\mI{{\bm{I}}}
\def\mJ{{\bm{J}}}

\def\mX{{\bm{X}}}

\def\mSigma{{\bm{\Sigma}}}

\DeclareMathAlphabet{\mathsfit}{\encodingdefault}{\sfdefault}{m}{sl}
\SetMathAlphabet{\mathsfit}{bold}{\encodingdefault}{\sfdefault}{bx}{n}
\newcommand{\tens}[1]{\bm{\mathsfit{#1}}}
\def\tA{{\tens{A}}}

\def\tX{{\tens{X}}}

\def\gG{{\mathcal{G}}}

\def\sA{{\mathbb{A}}}
\def\sB{{\mathbb{B}}}

\def\sS{{\mathbb{S}}}

\def\emA{{A}}

\newcommand{\etens}[1]{\mathsfit{#1}}

\def\etA{{\etens{A}}}

\newcommand{\E}{\mathbb{E}}

\newcommand{\R}{\mathbb{R}}

\newcommand{\KL}{D_{\mathrm{KL}}}
\newcommand{\Var}{\mathrm{Var}}

\newcommand{\Cov}{\mathrm{Cov}}

\newcommand{\normltwo}{L^2}
\newcommand{\normlp}{L^p}

\newcommand{\parents}{Pa} 

\DeclareMathOperator*{\argmax}{arg\,max}
\DeclareMathOperator*{\argmin}{arg\,min}

\usepackage{hyperref}
\usepackage{url}
\usepackage[misc]{ifsym}

\usepackage{graphicx}
\iclrfinalcopy 
\title{Joint Distribution-Informed Shapley Values for Sparse Counterfactual Explanations}

\author{%
Lei You\thanks{\Letter ~Corresponding Author.} \\
Technical University of Denmark \\
DK-2750 Ballerup, Denmark \\
\texttt{leiyo@dtu.dk} \\
\And
Yijun Bian \\
University of Copenhagen \\
DK-2100 Copenhagen, Denmark \\
\texttt{yibi@di.ku.dk} \\
\And 
Lele Cao \\
AI Labs, King (Microsoft Gaming) \\
SE-11157 Stockholm, Sweden \\
\texttt{lelecao@microsoft.com}
}

\usepackage[utf8]{inputenc} 
\usepackage[T1]{fontenc}    
\usepackage{hyperref}       
\usepackage{url}            

\usepackage{booktabs}       
\usepackage{nicefrac}       
\usepackage{microtype}      
\usepackage{mathtools}
\usepackage{amsmath,amssymb,amsthm}
\usepackage{bm}

\usepackage{enumerate}
\usepackage{enumitem}
\setlist{topsep=0pt, leftmargin=*}
\usepackage{xcolor}
\definecolor{darkblue}{rgb}{0.0, 0.0, 0.55}
\usepackage{float}
\usepackage{placeins} 
\usepackage{multirow}
\usepackage{tabularx}

\usepackage[hyperfirst=false, nogroupskip=true]{glossaries}
\glsdisablehyper
\usepackage{algorithm}
\usepackage{algorithmic}

\usepackage{mwe}
\usepackage{fontawesome}

\DeclareFontFamily{U}{mathx}{\hyphenchar\font45}
\DeclareFontShape{U}{mathx}{m}{n}{
      <5> <6> <7> <8> <9> <10>
      <10.95> <12> <14.4> <17.28> <20.74> <24.88>
      mathx10
      }{}
\DeclareSymbolFont{mathx}{U}{mathx}{m}{n}
\DeclareFontSubstitution{U}{mathx}{m}{n}
\DeclareMathAccent{\widebar}{0}{mathx}{"73}

\usepackage{appendix}
\let\appendixpagenameorig\appendixpagename
\renewcommand{\appendixpagename}{\normalfont\LARGE\scshape\appendixpagenameorig}
\usepackage{titletoc}
\usepackage{tocloft}
\usepackage{titlesec}

\usepackage{graphicx}
\usepackage{subcaption}
\usepackage{tikz}
\usepackage{pgfplots}
\pgfplotsset{compat=1.17}

\usepgfplotslibrary{fillbetween}
\usetikzlibrary{calc} 

\usepackage{array}
\usepackage[misc]{ifsym}

\theoremstyle{plain}
\newtheorem{theorem}{Theorem}[section]

\newtheorem{corollary}[theorem]{Corollary}
\theoremstyle{definition}

\theoremstyle{remark}

\providecolor{dtured}{rgb}{0.00, 0.00, 0.00} %

\newacronym{sr}{SR}{Sparse Regression}
\newacronym{ml}{ML}{machine learning}
\newacronym{xai}{XAI}{explainable artificial intelligence}
\newacronym{ot}{OT}{optimal transport}
\newacronym{eot}{EOT}{entropic optimal transport}
\newacronym{wd}{WD}{wasserstein distance}
\newacronym{mmd}{MMD}{maximum mean discrepancy}
\newacronym{meand}{MeanD}{absolute mean difference}
\newacronym{mediand}{MedianD}{absolute median difference}
\newacronym{ce}{CE}{counterfactual explanations}
\newacronym{fa}{FA}{feature attributions}
\newacronym{dce}{DCE}{distributional counterfactual explanation}
\newacronym{gce}{GCE}{group-based counterfactual explanation}
\newacronym{kl}{KL}{Kullback-Liebler}
\newacronym{clt}{CLT}{central limit theorem}
\newacronym{1d}{1D}{one-dimensional}
\newacronym{sgd}{SGD}{stochastic gradient descent}
\newacronym{bcd}{BCD}{block coordinated descent}
\newacronym{dkw}{DKW}{Dvoretzky-Kiefer-Wolfowitz}
\newacronym{ucl}{UCL}{Upper Confidence Limit}
\newacronym{mlpnet}{MLPNet}{multi-layer perceptron neural network}
\newacronym{svm}{SVM}{support vector machine}
\newacronym{rbfnet}{RBFNet}{radial basis function network}
\newacronym{bmi}{BMI}{body mass index}
\newacronym{asi}{ASI}{affine scale invariance}
\newacronym{dm}{DM}{demand monotonicity}
\newacronym{cola}{COLA}{{CO}unterfactuals with {L}imited {A}ctions}
\newacronym{p-shap}{$p$-SHAP}{joint-probability-informed Shapley}
\newacronym{milp}{MILP}{mixed integer linear programming}
\newacronym{scm}{SCM}{structural causal model}

\def\A{A}

\def\D{\mathcal{D}}
\def\F{\mathcal{F}}

\def\R{\mathcal{R}}
\def\RR{\mathbb{R}}

\def\EE{\mathbb{E}}

\def\diverg{D}

\hypersetup{
    colorlinks=true,
    linkcolor=darkblue,
    citecolor=darkblue,
    urlcolor=darkblue,
}

\usepackage[capitalise]{cleveref}

\DeclareMathAlphabet{\mathpzc}{OT1}{pzc}{m}{it}

\renewcommand{\vec}[1]{\mathbf{#1}}

\makeatletter
\let\oldmakefirstuc\makefirstuc
\renewcommand*{\makefirstuc}[1]{%
  \def\gls@add@space{}%
  \mfu@capitalisewords#1 \@nil\mfu@endcap
}
\def\mfu@capitalisewords#1 #2\mfu@endcap{%
  \def\mfu@cap@first{#1}%
  \def\mfu@cap@second{#2}%
  \gls@add@space
  \oldmakefirstuc{#1}%
  \def\gls@add@space{ }%
  \ifx\mfu@cap@second\@nnil
    \let\next@mfu@cap\mfu@noop
  \else
    \let\next@mfu@cap\mfu@capitalisewords
  \fi
  \next@mfu@cap#2\mfu@endcap
}
\makeatother

\renewcommand{\eqref}[1]{equation~(\ref{#1})}

\begin{document}

\maketitle

\begin{abstract}
Counterfactual explanations (CE) aim to reveal how small input changes flip a model’s prediction, yet many methods modify more features than necessary, reducing clarity and actionability. We introduce \emph{COLA}, a model- and generator-agnostic post-hoc framework that refines any given CE by computing a coupling via optimal transport (OT) between factual and counterfactual sets and using it to drive a Shapley-based attribution (\emph{$p$-SHAP}) that selects a minimal set of edits while preserving the target effect. Theoretically, OT minimizes an upper bound on the $W_1$ divergence between factual and counterfactual outcomes and that, under mild conditions, refined counterfactuals are guaranteed not to move farther from the factuals than the originals. Empirically, across four datasets, twelve models, and five CE generators, COLA achieves the same target effects with only 26--45\% of the original feature edits. On a small-scale benchmark, COLA shows near-optimality. 
\vspace{0.05cm}

\noindent \textbf{Experiment code}: {\small \url{https://github.com/youlei202/XAI-COLA}}.

\noindent \textbf{Software} \citep{zhu2026xaicola}: {\small \url{https://pypi.org/project/xai-cola/}}. 
\end{abstract}

\section{Background}
\label{sec:background}

\Gls{xai} is essential for making artificial intelligence systems transparent and trustworthy \citep{arrieta2020explainable,das2020opportunities}. Within this area, \gls{fa} methods, such as Shapley values \citep{sundararajan2020many, lundberg2017unified}, determine how much each input feature contributes to a \gls{ml} model's output. This helps simplify complex models by highlighting the most influential features. For example, in a healthcare model, Shapley values can identify key factors like age and medical history, assisting clinicians in understanding the model's decisions \citep{ter2023challenges,nohara2022explanation}. Another technique \gls{ce} \citep{wachter2017counterfactual,guidotti2022counterfactual} show how small changes in input features can lead to different outcomes. While hundreds of \gls{ce} algorithms have been proposed \citep{guidotti2022counterfactual, verma2020counterfactual} to date, it is hardly practical to find one single \gls{ce} algorithm that suits for all user cases, due to each of them is tailored particularly for their own different scenarios, goals, and tasks. For instance, the objective in one \gls{ce} algorithm can be defined as finding a single counterfactual instance for each factual instance sometimes, while at othertimes, it could be treating multiple instances as a group and seeking one or more/multiple counterfactual instances for each factual observation. In some cases, the focus of a \gls{ce} algorithm could be on the entire dataset, aiming to identify global \gls{ce} that indicate the direction to move the factual instances to achieve the desired model output \citep{rawal2020beyond,ley2022global,ley2023globe,carrizosa2024mathematical}. Yet in other scenarios, the group of factual instances is viewed as a distribution, aiming to find a counterfactual distribution that remains similar in shape to the factual distribution \citep{you2024discount}, and ensuring comparable costs. Besides, some \gls{ce} algorithms assume differentiable models, whereas others are designed specifically for tree-based or ensemble models.


\paragraph{Problem Description and Challenges} 
We address a general and comprehensive problem in this paper, which builds on the extensive foundations established in the literature (see Appendix~\ref{subsec:related_work}). To be specific, we seek to answer the following question:
Although both \gls{fa} and \gls{ce} are vital for making AI models more interpretable and accountable, relying only on one of them will cause drawbacks. That is, \gls{fa} alone may not provide tangible steps as explanations, and CE alone might include unnecessary feature changes that are not practical. Therefore, we address a general and comprehensive problem in this paper, which builds on the extensive foundations established in the literature (see Appendix~\ref{subsec:related_work}). To be specific, we seek to answer the following question:
%
\begin{figure}[h]
\centering
\includegraphics[width=.867\linewidth]{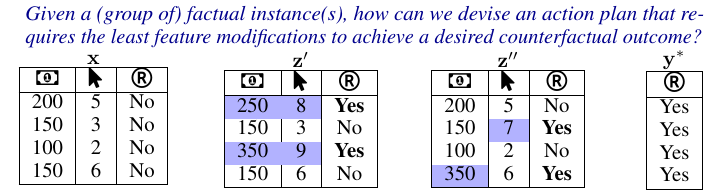}
\caption{[\textit{Example: User engagement on an e-commerce platform}] A platform aims to increase user registrations. The platform has collected data on user interactions, such as the amount of money spent (\faMoney), the number of clicks (\faMousePointer), and whether the user has registered (\faRegistered). In the original data ($\vec{x}$), no users are registered. Action plans $\vec{z}'$ and $\vec{z}''$ adjust user characteristics to achieve the desired outcome ($\vec{y}^*$) of full registration. Both plans achieve a half counterfactual effect, but $\vec{z}''$ requires fewer modifications compared to $\vec{z}'$. 
}\label{fig:problem}
\end{figure}

Three major challenges remain in addressing this problem. First, it is unrealistic to expect a single \gls{ce} algorithm to meet all the needs universally, as the problem is often task-specific. Second, the approach should not rely on strong assumptions about the model (for example, requiring differentiability or special structures) to ensure its applicability across a wide range of models. Third, \gls{fa} like feature importance can be misleading due to the lack of coherence between the \gls{fa} scores and the changes for counterfactual effect. In other words, it is not effective to \textcolor{dtured}{perform \gls{fa} independently of \gls{ce} to select the most important features to change.} We will demonstrate later
 that this decoupling can result in counterproductive feature modifications (also referred to as actions from users/stakeholders) \textcolor{dtured}{in Result II of Section~\ref{sec:numerical}}, as the features deemed important generally may not align with the specific pathways to achieve the desired counterfactual outcomes.

\paragraph{Main Contributions} We introduce \gls{cola}, a general post-hoc framework that refines \gls{ce} across models and \gls{ce} generators by using an \gls{ot}–induced coupling between factual and counterfactual sets to guide Shapley attributions. This yields $p$-SHAP, which unifies other commonly used Shapley methods under appropriate couplings and offers a modular interface for attribution and edit selection. We provide theoretical guarantees: \gls{ot} minimizes an upper bound on the $W_1$ divergence, and under mild conditions refined counterfactuals are provably no farther from factuals than the originals. Empirically, across four datasets, twelve models, and five \gls{ce} methods, \gls{cola} with $p$-SHAP achieves the same target effects with only 26–45\% of the original feature edits. In a small-scale benchmark, \gls{cola} is near-optimal.


 To our best knowledge, this is the first method proposed for systematically addressing the problem in \figurename~\ref{fig:problem} \textit{without specifying certain \gls{ce} algorithms and models}.

\section{Problem Formulation}
We formally formulate the problem described in \figurename~\ref{fig:problem}. Denote $f:\mathbb{R}^d \rightarrow \mathbb{R}$ as a black-box \gls{ml} model. Denote by $\vec{x}$ any observed (factual) data with $n$ rows and $d$ columns ($\vec{x}\in\RR^{n\times d}$, $n\geq 1$, and $d\geq 1$). Let $\vec{y}^*$ be the target model output ($\vec{y}^*\in\RR^{m}$, $m\geq 1$). 
The optimization is to look for a (group of) counterfactual data instance(s) $\vec{z}$ ($\vec{z}\in\RR^{n\times d}$) subject to a maximum number of allowed feature changes, $C$, to achieve model output(s) as close as possible to $\vec{y}^*$. Let $D$ denote a divergence function that measures the dissimilarity between two entities. The problem is below.
\begin{subequations}
\begin{align} 
\min_{\vec{c},\vec{z}} \quad & \diverg\left(f(\vec{z}), \vec{y}^*\right)  \,,\label{eq:main_problem_obj} \\ 
\text{s.t.} \quad & \diverg\left(\vec{z},\vec{x}\right) \leq \epsilon \,,\label{eq:main_problem_epsilon}\\
                  & \textstyle \sum_{i=1}^n\sum_{k=1}^d c_{ik} \leq C \,,\label{eq:main_problem_C}\\
                  & z_{ik} \leq x_{ik} (1-c_{ik}) + M c_{ik} \quad i=1,\ldots, n\,,~\ k=1,\ldots, d \,,\label{eq:main_problem_constr} \\
                  & z_{ik} \geq x_{ik} (1-c_{ik}) - M c_{ik} \quad i=1,\ldots, n\,,~\ k=1,\ldots, d \,.\label{eq:main_problem_constr_2} 
\end{align}
\label{eq:main_problem}%
\end{subequations}
The objective \eqref{eq:main_problem_obj} and the constraint \eqref{eq:main_problem_epsilon} formulate the typical \gls{ce} optimization. Namely, $\vec{z}$ is expected to make $f(\vec{z})$ close to $\vec{y}^*$ yet stays close to $\vec{x}$. Then $\vec{z}$ can be used as a counterpart reference to explain why $f(\vec{x})$ does not achieve $\vec{y}^*$. We do not limit the function $D$ to any specific type of divergence function, allowing it to stay general. Example functions of $D$ can be Euclidean distance, \gls{ot}, \gls{mmd}, or differences of the model outcome in mean or median. Then, \eqref{eq:main_problem_C}--\eqref{eq:main_problem_constr_2} compose the \gls{ce} optimization constrained by  actions. On top of the counterfactual data $\vec{z}$, we also optimize an indicator variable $\vec{c}$, such that $z_{ik}$ is not allowed to change iff $c_{ik}=0$. Maximum $C$ changes are allowed as imposed by \eqref{eq:main_problem_C}. Inspecting \eqref{eq:main_problem_constr} and \eqref{eq:main_problem_constr_2}, if $c_{ik}=0$, $x_{ik}$ equals $z_{ik}$ and no changes happen at ($i,k$). Otherwise, if $c_{ik}=1$, remark that $M$ is a sufficiently large constant such that $z_{ik}$ has good freedom to change.

\textcolor{dtured}{To solve \eqref{eq:main_problem}, we resort to \gls{fa} to identify the most influential features to obtain the modification indicator variable $\vec{c}$. The next section introduces commonly used Shapley value methods for \gls{fa}, which, together with our later proposed one, are integrated into our algorithmic framework COLA, to obtain the refined counterfactual $\vec{z}$.} The problem is computationally difficult even when $d=1$ for linear models, see Appendix~\ref{subsec:np-hard}.

\section{Preliminaries on Shapley Value}
\label{sec:baseline_shapley}

This section introduces commonly used Shapley value methods for \gls{fa}, which, together with our later proposed one, are integrated into our algorithmic framework \gls{cola}. We first introduce the concept of Shapley value in game theory, followed by a discussion on various commonly used Shapley methods for \gls{fa}. 
Readers could find in \citep{sundararajan2020many} for a comprehensive overview of different Shapley value methods.

 In cooperative game theory, a coalitional game is characterized by a finite set of players (in our context, features), denoted by \( \F=\{1,2,\ldots, d\} \), and a characteristic function \( v \). This function \( v: 2^\F \rightarrow \mathbb{R} \) maps each subset \( S \subseteq \F \) to a real number \( v(S) \), representing the total payoff that can be achieved by the members of \( S \) through cooperation, with the condition \( v(\emptyset) = 0 \). The Shapley value, as defined based on $v$ later, is a fundamental concept in this framework, providing an equitable method to distribute the overall payoff, \( v(\F) \), of the grand coalition among the individual players.

Prior to defining the Shapley value, denote \( \Delta(k, S) \) the incremental value produced by a player (feature) \( k \) to a coalition \( S \), aligning with the concept of marginal contribution. This is defined as the additional payoff realized by the coalition due to the inclusion of the player \( k \), i.e.,
\begin{equation}
    \Delta(k, S)=v(S\cup k) - v(S).
\end{equation}

The Shapley value \( \phi_i \) for a player \( i \) in a cooperative game is defined in \eqref{eq:phi} below.
\begin{equation}
    \phi_k =\frac{1}{d} \sum_{S\subset \F\backslash\{k\}} \begin{pmatrix} d-1 \\ |S|\end{pmatrix}^{-1}\Delta(k, S).
\label{eq:phi}
\end{equation}

The value $\phi_k$ is computed by averaging the normalized marginal contributions \( \Delta(k, S) \) of feature \( k \) across all subsets \( S \) of the feature set excluding \( k \). The normalization factor is the inverse of the binomial coefficient \( \binom{d-1}{|S|} \), where \( d \) is the total number of features. By summing these values for all subsets and dividing by \( d \), the equation provides a fair and axiomatic distribution of payoffs among players, reflecting each features's contribution among the whole collection of features. 

For an arbitrary data point $\vec{x}_i$, we use $\vec{x}_{i,S}$ to represent its part containing only the features in $S$, and $\vec{x}_{i,\F\backslash S}$ the other remaining part. The same rule applies for other notations as well. Below, we introduce commonly used Shapley value methods.

\textit{Baseline Shapley (B-SHAP)} . This approach relies on a specific baseline data point $\vec{r}_j$ as the reference for any $\vec{x}_i$. The set function is defined to be 
\begin{equation}
v^{(i)}_{\text{B}}(S)=f(\vec{x}_{i,S};\vec{r}_{j,\F\backslash S}) - f(\vec{r}_j),
\label{eq:v_B}
\end{equation}
where a feature's absence is modeled using its value in the reference baseline data point $\vec{r}$ \citep{lundberg2017unified,sun2011axiomatic,merrick2020explanation}. This method assumes an exact alignment between $\vec{x}$ and $\vec{r}$.

\textit{Random Baseline Shapley (RB-SHAP)}. This approach is a variant of BShap and is implicitly used in \citep{lundberg2017unified} as well as the famous ``SHAP'' library. The equation is as follows.
\begin{equation}
    v^{(i)}_{\text{RB}}(S) = \EE_{\vec{x}'\sim\D}\left[f(\vec{x}_{i,S};\vec{x}'_{\F\backslash S })\right] - \EE_{\vec{x}'\sim \D}[f(\vec{x}')].
\label{eq:v_RB}
\end{equation}
where $\D$ is the background distribution that is commonly selected to be the training dataset \citep{lundberg2017unified,merrick2020explanation}. 

\textit{Counterfactual Shapley (CF-SHAP)}. This is an extension from RB-SHAP, by taking the distribution $\D$ as the counterfactual distribution defined on every single factual instance $\vec{x}_i$ ($i=1,2,\ldots, n$).
\begin{equation}
v^{(i)}_{\text{CF}}(S) = \EE_{\vec{r}\sim\D(\vec{x}_i)}\left[f(\vec{x}_{i,S};\vec{r}_{\F\backslash S })\right] - \EE_{\vec{r}\sim \D(\vec{x}_i)}[f(\vec{r})].
\label{eq:v_CF}
\end{equation}
This method has been adopted in \citep{albini2022counterfactual,kommiya2021towards}, demonstrated with advantages for contrastive \gls{fa}. This method assumes a probabilistic alignment between $\vec{x}$ and $\vec{r}$.

\section{The Proposed \gls{p-shap} and Its Theoretical Aspects}

\paragraph{(\textcolor{darkblue}{Proposed}) \Gls{p-shap}} We generalize \eqref{eq:v_B} to \eqref{eq:v_CF} by integrating an algorithm $\A_{\text{Prob}}$ that returns their joint probability.\footnote{First, \gls{p-shap} degrades to \textcolor{dtured}{B-SHAP} in \eqref{eq:v_B} when $\A_{\text{Prob}}$ defines \textcolor{dtured}{a joint distribution between $\vec{x}$ and $\vec{r}$ that indicates} an $i\leftrightarrow j$ alignment of for any $\vec{x}_i$, $\vec{r}_j$.
Second, \gls{p-shap} degrades to RB-SHAP in \eqref{eq:v_RB} when  $\A_{\text{Prob}}$ is defined to be independent of \gls{ce} but associates with an arbitrary distribution $\D$.
Third, \gls{p-shap} degrades to CF-SHAP in \eqref{eq:v_CF} when $\A_{\text{Prob}}$ is built upon a known distribution of \gls{ce}. } Our \gls{p-shap} is defined as follows,
\begin{subequations}
\begin{align}
    v^{(i)}(S) = & ~\EE_{\vec{r}\sim p(\vec{r}|\vec{x}_i)}\left[f(\vec{x}_{i,S}; \vec{r}_{\F\backslash S})\right] - \EE_{\vec{r}\sim p(\vec{r})}\left[f(\vec{r})\right] \,,\\
    \text{s.t.}&\quad   p = \A_{\text{Prob}}(\vec{x},\vec{r}) \,.
\end{align}
    \label{eq:v_f}%
\end{subequations}
\textcolor{black}{Crucially, unlike prior methods that rely on random baselines, p-SHAP solves the alignment problem by identifying the optimal coupling via OT. This reformulates feature attribution not merely as a prediction decomposition, but as a transport problem that minimizes the cost of explanation.}

Remark that $\A_{\text{Prob}}$ \textit{is independent of the \gls{ce} algorithm, but only depends on the generated counterfactual $\vec{r}$.}
By focusing solely on these fixed components, $p$-SHAP ensures consistency in \gls{fa} without being influenced by the variability of different \gls{ce} generation processes, which is a major difference to CF-SHAP.
Contrary to common expectations, we demonstrate \textcolor{dtured}{that \gls{ot} can be more effective than relying on a counterfactual distribution defined by a \gls{ce} generation mechanism as done by \cite{albini2022counterfactual}, in Result II of Section~\ref{sec:numerical} later.}

Especially, one of the focus in this paper is to consider the \gls{ot} problem (also the $2$-Wasserstein divergence) defined below. And the transportation plan $\vec{p}_{\text{OT}}$ obtained by solving \gls{ot} is used as the joint distribution of $\vec{x}$ and $\vec{r}$ in \gls{p-shap}.
\begin{equation}
\vec{p}_{\text{OT}} = \argmin_{\vec{p} \in \Pi(\bm{\mu}, \bm{\nu})} \sum_{i=1}^n \sum_{j=1}^m p_{ij} \|\vec{x}_i - \vec{r}_j\|_2^2 + \varepsilon \sum_{i=1}^n \sum_{j=1}^m p_{ij} \log \left( \frac{p_{ij}}{\mu_i \nu_j} \right).
\label{eq:ot}
\end{equation}

Note that $\bm{\mu}$ and $\bm{\nu}$ represent the marginal distributions of $\vec{x}$ and $\vec{r}$ respectively,  and 
\( \Pi(\bm{\mu}, \bm{\nu}) \) the set of joint distributions (i.e. all possible transport plans).  The term \( \varepsilon \sum_{i=1}^n \sum_{j=1}^m p_{ij} \log ( {p_{ij}}/{(\mu_i \nu_j)} ) \) is the entropic regularization with \( \varepsilon\geq 0 \) being the coefficient. Such regularization ($\varepsilon>0$) helps accelerate the computation of \gls{ot}.

\paragraph{Theoretical Aspects of \gls{p-shap}} 
\gls{ot} minimizes the total feature modification cost (i.e. modifying $\vec{x}$ towards $\vec{r}$) under its obtained alignment between factual $\vec{x}$ and counterfactual $\vec{r}$. This directly corresponds to our objective of finding feature modifications that achieve the counterfactual outcomes at minimal cost. 
We can further strengthen this connection theoretically under the Lipschitz continuity assumption of the predictive model $f$. In Theorem \ref{thm:lipschitz} below (proof in Appendix \ref{subsec:ot_theory}),  we establish that the transportation plan $\vec{p}_{\text{OT}}$ used in \gls{p-shap} is effective in minimizing an upper bound on the divergence between $f(\vec{x})$ and $\vec{y}^*$. Specifically, the $1$-Wasserstein distance between $f(\vec{x})$ and $\vec{y}^*$, is bounded by the Lipschitz constant (assuming Lipschitz continuity of $f$) multiplied by the square root of the minimized expected cost of changing $\vec{x}$ towards $\vec{r}$, i.e. $\sum_{i,j}p_{ij}\|\vec{x}_i-\vec{r}_j\|_2^2$ where \( p_{ij} \) (\( j = 1, 2, \ldots, m \)) quantify how the feature values of \( \vec{x}_i \) should be adjusted towards those of one or multiple \( \vec{r}_j \).
Practically, this means that in \gls{p-shap}, the \gls{ot} plan \( \vec{p}_{\text{OT}} \) provides a strategy to adjust the feature values of \( \vec{x} \) towards those of \( \vec{r} \) in a way that minimizes the expected modification cost \( \sum_{i,j} p_{ij} \| \vec{x}_i - \vec{r}_j \|_2^2 \). Compared to other modification plans ($\vec{p}\in\Pi$), $\vec{p}_{\text{OT}}$ yields the minimal possible cost, which in turn provides the tightest upper bound on the violation of the counterfactual effect \( W_1(f(\vec{x}), \vec{y}^*) \) in proportion to this cost.

\begin{theorem}[\gls{p-shap} Towards Counterfactual Effect] Consider the $1$-Wasserstein divergence $W_1$, i.e.
\(
W_1(f(\vec{x}),\vec{y}^*)=\min_{\bm{\pi}\in\Pi}\sum_{i=1}^n \sum_{j=1}^m \pi_{ij} \left| f(\vec{x}_i) - \vec{y}^*_j \right|
\).
Suppose the counterfactual outcome $\vec{y}^*$ is fully achieved by $\vec{r}$, i.e. $\vec{y}^*_j = f(\vec{r}_j)$ ($j=1,2,\ldots, m$).
Assume that the model \( f: \mathbb{R}^d \rightarrow \mathbb{R} \) is Lipschitz continuous with Lipschitz constant \( L \).
The expected absolute difference in model outputs between the factual and counterfactual instances, weighted by \( \vec{p}_{\text{OT}} \) (with $\varepsilon=0$), is bounded by:
\[
W_1(f(\vec{x}),\vec{y}^*) \leq L \sqrt{ \sum_{i=1}^n \sum_{j=1}^m p_{ij}^{\text{OT}} \| \vec{x}_i - \vec{r}_j \|_2^2 }\leq L \sqrt{ \sum_{i=1}^n \sum_{j=1}^m p_{ij} \| \vec{x}_i - \vec{r}_j \|_2^2 } 
\,,\quad \forall \vec{p}\in\Pi \,.
\]
Namely, $\vec{p}_{\text{OT}}$ minimizes the upper bound of $W_1(f(\vec{x}),\vec{y}^*)$, where the upper bound is based on the expected feature modification cost.
\label{thm:lipschitz}
\end{theorem}

\textcolor{black}{Although Theorem~\ref{thm:lipschitz} bounds the transport cost, it serves as a convex proxy for the $\mathcal{NP}$-hard discrete sparsity ($L_0$) problem. By concentrating attribution mass on the most efficient paths, the OT coupling effectively guides the greedy selection in Algorithm~\ref{alg:cola} to discard non-essential features.}
In addition, \gls{p-shap} is conceptually correct in attributing the causal behavior to the modifications of the characteristics, stated in Theorem~\ref{thm:intervention} below (proof in Appendix \ref{sec:proof_thm_intervention}).
\begin{theorem}[Interventional Effect of \gls{p-shap}]
For any subset \( S \subseteq \F \) and any \( \vec{x}_i \) (\(i=1,2,\ldots,n\)), \( v^{(i)}(S) \) represents the causal effect of the difference between the expected value of \( f(\vec{r}) \) under the intervention on features \( S \) and the unconditional expected value of \( f(\vec{r}) \). Mathematically, this is expressed as:
\[
\EE[f(\vec{r})]+v^{(i)}(S) =  \EE\left[f(\vec{r}) \middle| do\left(\vec{r}_S = \vec{x}_{i,S}\right) \right]. 
\]
\label{thm:intervention}
\end{theorem}
Furthermore, we remark that \gls{p-shap} preserves nice axioms of B-SHAP and RB-SHAP, which makes it an effective tool for attributing features. We omit the proof but refer to \citep{sundararajan2020many,lundberg2017unified} as a reference for axioms of Shapley.

\section{The Algorithmic Framework \gls{cola}}

\begin{figure}[t]
\includegraphics[width=1.01\linewidth]{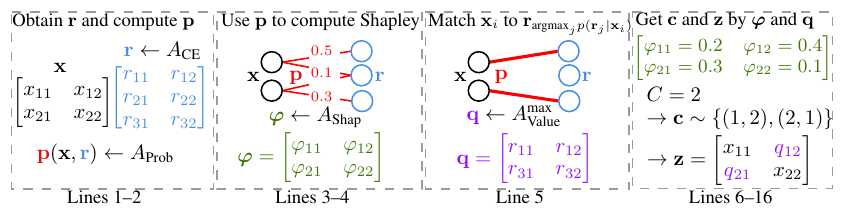}
\caption{[\textit{An illustration of COLA}] This figure shows how COLA gets $\vec{c}$ and $\vec{z}$ for  \eqref{eq:main_problem}. We use $\A^{\text{max}}_{\text{Value}}$ for illustration in line~\ref{alg:cola-A_value} due to its simplicity. In lines 6--16, we assume $C=2$, and the sampling yields exactly two positions for modfications according to the probability matrix $\bm{\varphi}$. 
}\label{fig:cola}
\end{figure}

\paragraph{Sketch} The algorithmic framework \gls{cola}, stated in Algorithm~\ref{alg:cola} below, aims to solve \eqref{eq:main_problem} and is established on four categories of algorithms.
 We explain Algorithm~\ref{alg:cola} in details below, along with an illustration in \figurename~\ref{fig:cola}.

\begin{algorithm}[h]
\caption{{CO}unterfactuals with {L}imited {A}ctions ({COLA})}
\label{alg:find_max}
\begin{algorithmic}[1]
\REQUIRE Model $f$, factual $\vec{x}\in\RR^{n\times d}$, target $\vec{y}^*\in\RR^{m}$, $\epsilon$, and $C$
\ENSURE Action plan $\vec{c}\in\RR^{n\times d}$ and correspondingly a refined counterfactual $\vec{z}\in\RR^{n\times d}$
\STATE Use $\A_{\text{CE}}(f, \vec{x}, \vec{y}^*, \epsilon)$ to obtain  $\vec{r}\in\RR^{m\times d}$, with $\vec{y}^{*}= f(\vec{r})$ and $D(\vec{r}, \vec{x})\leq \epsilon$. \label{alg:cola-A_ce}
\STATE Use $\A_{\text{Prob}}(\vec{x},\vec{r})$ to obtain the joint distribution matrix $\vec{p}\in\RR_{+}^{n\times m}$. \label{alg:cola-A_prob}
\STATE Use $\A_{\text{Shap}}(\vec{x}, \vec{r}, \vec{p})$ to obtain the shapley values $\bm{\phi}\in\RR^{n\times d}$ for each element of $\vec{x}$. \label{alg:cola-A_shap}
\STATE Normalize the element-wise absolute values of $\bm{\phi}$, i.e., $\varphi_{ik}\gets |\phi_{ik}|/\|\bm{\phi}\|_1$ ($\bm{\varphi}\in\RR_{+}^{n\times d}$). \label{alkg:cola-phi}
\STATE Use $\A_{\text{Value}}(\vec{r},\vec{p})$ to obtain matrix $\vec{q}\in\RR^{n\times d}$. \label{alg:cola-A_value}
\STATE For $\vec{c}\in\{0,1\}^{n\times d}$, $c_{ik}\gets 0$ $\,(i=1,\ldots, n\,,~\ k=1,\ldots, d)$. \label{alg:cola-c-begin}
\STATE Sample $C$ pairs ($i$, $k$) according to the probability matrix $\bm{\varphi}$, and let $c_{ik}=1$ for them. \label{alg:cola-c_ik}
\STATE Let $\vec{z}\gets \vec{x}$ ($\vec{z}\in\RR^{n\times d}$).
\FOR{$i \gets 1$ to $n$}
    \FOR{$k \gets 1$ to $d$}
        \IF{$c_{ik}=1$}
            \STATE $z_{ik}\gets q_{ik}$
        \ENDIF
    \ENDFOR
\ENDFOR
\RETURN $\vec{c}$ and $\vec{z}$ \label{alg:cola-c-end}
\end{algorithmic}
\label{alg:cola}
\end{algorithm}
%

\textbf{Line~\ref{alg:cola-A_ce}} (\textit{Applying a \gls{ce} algorithm to find a counterfactual $\vec{r}$}). The \gls{ce} algorithm $\A_{\text{CE}}$ takes the model $f$, the factual $\vec{x}$, the target outcome $\vec{y}^*$, and the tolerance $\epsilon$ as input. The algorithm returns a counterfactual  $\vec{r}$ staying close with $\vec{x}$, with $\vec{y}^{*}= f(\vec{r})$ and $D(\vec{r},\vec{x})\leq \epsilon$. 

\textbf{Line~\ref{alg:cola-A_prob}} (\textit{Seeking a joint distribution of $\vec{x}$ and $\vec{r}$}). We use an algorithm $\A_{\text{Prob}}$ for this task, which takes $\vec{x}$ and $\vec{r}$ as input, and outputs a matrix representing the joint distribution of all $n$ and $m$ data points in $\vec{x}$ and $\vec{r}$, respectively. The joint distribution $\vec{p}$ represents an alignment relationship (or matching) between the factual rows and counterfactual rows, and we use it in Line~\ref{alg:cola-A_value} to compute the values that can be used for composing $\vec{z}$ later on. As discussed in Section~\ref{sec:baseline_shapley}, $\A_{\text{Prob}}$ can be based on \gls{ot} to compute a joint distribution that yields the smallest \gls{ot} distance between $\vec{x}$ and $\vec{r}$, if the alignment relationship between their rows are unknown. Otherwise, it is recommended to select a joint distribution that accurately reflects the alignment between the rows in $\vec{x}$ and $\vec{r}$.

\textbf{Lines~\ref{alg:cola-A_shap}--\ref{alkg:cola-phi}} (\textit{\gls{p-shap} \gls{fa}}). 
We apply \eqref{eq:v_f}  to compute the shapley value for $\vec{x}$. The joint distribution $\vec{p}$ can be used here (without being forced) to properly align each row of $\vec{x}$ with its corresponding counterfactual rows of $\vec{r}$, such that the selected rows in $\vec{r}$ serve as the most representative contrastive reference for the row in $\vec{x}$. Numerically, this alignment significantly influences our contrastive \gls{fa}. 
Then, the shapley values (as a matrix) of $\vec{x}$ is taken element-wisely with the absolute values and normalized (such that all values sum up to one). The resulted matrix $\bm{\varphi}$ forms our \gls{fa}.

\textbf{Line~\ref{alg:cola-A_value}} (\textit{Computing feature values}). The algorithm $\A_{\text{Value}}$ is used for this task, which takes the counterfactual $\vec{r}$ and the joint distribution $\vec{p}$ as input. For any row $i$ in $\vec{x}$,  $\A_{\text{Value}}$ 
selects one or multiple row(s) in $\vec{r}$, used as references for making changes in $\vec{x}$.
 The algorithm returns a matrix $\vec{q}\in\RR^{n\times d}$, where each element $q_{ik}$ serves as a counterfactual candidate for $x_{ik}$ ($\forall i,k$). Below, we introduce $\A^{\text{max}}_{\text{Value}}$ and $\A^{\text{avg}}_{\text{Value}}$, respectively for the cases of selecting single row and selecting multiple rows.

For any row $\vec{x}_i$,  $\A^{\text{max}}_{\text{Value}}$ selects the row of $\vec{r}$ with the highest probability.
\begin{equation}
\vec{q} = \A^{\text{max}}_{\text{Value}}(\vec{r},\vec{p}) \,,\;
\text{ where $q_{ik}=r_{\tau(i),k}$ and $\tau(i) = \argmax_{j=1,2,\ldots, m}p_{ij}$}
\,. \label{eq:A_value_max}
\end{equation}
The algorithm $\A^{\text{avg}}$ computes $q_{ik}$ as a convex combination as a weighted average of $r_{1k}, r_{2k}, \ldots r_{mk}$. 
\begin{equation}
\vec{q} = \A^{\text{avg}}_{\text{Value}}(\vec{r},\vec{p}) \,,\;
\text{ where } q_{ik}=\sum_{j=1}^m\left(\frac{p_{ij}}{\sum_{j'=1}^m p_{ij'}}\right)r_{jk}
\,. \label{eq:A_value_avg}
\end{equation}
\textbf{Lines~\ref{alg:cola-c-begin}--\ref{alg:cola-c-end}}. Recall that the non-negative matrix $\bm{\varphi}$ is normalized to have its summation being one. We could hence treat it as a policy to select the positions in $\vec{x}$ for value replacement (i.e. $c_{ik}=1$), as what line~\ref{alg:cola-c_ik} does. Then, for any $i$ and $k$ with $c_{ik}=1$, $x_{ik}$ gets modified to $q_{ik}$, and the modified matrix is then returned as $\vec{z}$ together with $\vec{c}$ forming the optimized solutions of the problem in \eqref{eq:main_problem}.

\textcolor{dtured}{By Theorem~\ref{thm:lipschitz}, COLA is designed to minimize the dissimilarity between $f(\vec{z})$ and $\vec{y}^*$ by modifying $\vec{z}$ based on feature attribution results, which identify the most important features to adjust to achieve the desired counterfactual effect.}
The theorem below (proof in Appendix~\ref{sec:z_distance_proof}) demonstrates that the refined \( \vec{z} \), produced by the \gls{cola} framework, satisfies the constraint \eqref{eq:main_problem_epsilon} in the typical scenario where \( n = m \), using the Frobenius norm as the distance measure. Empirical evidence supporting the general applicability of this conclusion can be found in \tablename~\ref{tab:features}.
\begin{theorem}[Counterfactual Proximity]
\label{thm:z_distance}
\textcolor{black}{Let \( \vec{q} \in \mathbb{R}^{n \times d} \) be the aligned reference matrix derived from the counterfactual set \( \vec{r} \in \mathbb{R}^{m \times d} \) via the value assignment step \( \vec{q} = A_{\text{Value}}(\vec{r}, \vec{p}) \), where \( \vec{p} \) is the coupling matrix. For any dimensions \( n, m \ge 1 \), the refined counterfactual \( \vec{z} \) constructed by the COLA framework satisfies:
\[
\| \vec{z} - \vec{x} \|_F \leq \| \vec{q} - \vec{x} \|_F \,.
\]
This inequality guarantees that the refined counterfactual \( \vec{z} \) is strictly closer (or equal) to the factual input \( \vec{x} \) compared to the aligned counterfactual proposal \( \vec{q} \), ensuring that the refinement process introduces no additional divergence.
}
\end{theorem}

\begin{corollary}
\label{cor:z_distance_permutation}
\textcolor{black}{
In the special case where \( n = m \) and the OT plan \( \vec{p} \) corresponds to a deterministic permutation \( \sigma \) (i.e., obtained without entropic regularization, \( \varepsilon = 0 \)), the matrix \( \vec{q} \) becomes a row-permuted version of \( \vec{r} \) (denoted as \( \vec{r}_{\sigma} \), where \( q_{i} = r_{\sigma(i)} \)). In this setting, the bound simplifies to:
\[
\| \vec{z} - \vec{x} \|_F \leq \| \vec{r}_{\sigma} - \vec{x} \|_F \,,
\]
recovering the intuition that the refined counterfactual is at least as close to the factuals as the original counterfactual set reordered by \( \sigma \).
}
\end{corollary}

\paragraph{Complexity of \gls{cola}} Let $\mathcal{O}(M_{\text{CE}})$ be the algorithm complexity of $\A_{\text{CE}}$. For algorithm $\A_{\text{Shap}}$, consider using weighted linear regression to estimate Shapley values, and denote by $M_{\text{Shap}}$ the number of sampled subsets. The complexity of \gls{cola} with respect to $n$, $m$, $d$, and the regularization parameter $\varepsilon$ of entropic \gls{ot} is 
\(
\mathcal{O}(M_{\text{CE}}) + \mathcal{O}(n m \log(1/\varepsilon)) + \mathcal{O}(n d M_{\text{Shap}}) + N
\)
where $N=\mathcal{O}(nm) + \mathcal{O}(nd)$ if $\A^{\text{max}}_{\text{Value}}$ is used and $N=\mathcal{O}(nmd)$ if $\A^{\text{avg}}_{\text{Value}}$ is used. See Appendix \ref{subsec:complexity} for more details.

\section{Numerical Results}
\label{sec:numerical}
This section evaluates the effectiveness of \gls{cola} in addressing the problem in \eqref{eq:main_problem}, with $\vec{y}^{*}=f(\vec{r})$ where $\vec{r}$ is the counterfactual obtained from a \gls{ce} method $\A_{\text{CE}}$.  We adopt four different divergence functions: \Gls{ot} evaluates the distance between entire distributions. \Gls{mmd}  evaluates the divergence between the means of two distributions in a high-dimensional feature space. The \gls{meand} and \gls{mediand} evaluate the divergence between mean and median, respectively. The numerical results aim at showing:
I) \textit{\gls{cola}'s effectiveness for modification minimality.}
II) \textit{\gls{p-shap}'s superior performance than other Shapley methods towards modification minimality.}
III) \textit{\gls{cola}'s near-optimal performance.}

\paragraph{Experiment Setup} 
The experiments\footnote{The code is available on both \faGithub \\  \url{https://anonymous.4open.science/r/Contrastive-Feature-Attribution-DFB1} and the submitted supplementary files.} are conducted with $4$ datasets for binary classification tasks, $5$ \gls{ce} algorithms that are designed for diverse goals, and $12$ classifiers, shown in \tablename~\ref{tab:experiment-setup}, where a combination of dataset, $A_{\text{CE}}$ algorithm, and a model defines an ``experiment scenario''.

\begin{table}[h!]
\centering
\caption{[\textit{Experiment Scenarios Setup}] Four datasets are used to benchmark five \gls{ce} algorithms over $12$ models. A ``scenario'' is defined to be a combination of dataset, $A_{\text{CE}}$ algorithm, and a model $f$.}%
\vspace{-2mm}
\setlength{\extrarowheight}{2pt}
\begin{tabular}{p{1.35cm}|p{11.8cm}}
\toprule
\textbf{Dataset} & HELOC \citep{holter2018fico}, German Credit \citep{misc_statlog_(german_credit_data)_144}, Hotel Bookings \citep{antonio2019hotel}, COMPAS \citep{jeff2016we} \\ \hline
$\A_{\text{CE}}$ & DiCE \citep{mothilal2020explaining}, AReS \citep{rawal2020beyond}, GlobeCE \citep{ley2023globe}, KNN \citep{albini2022counterfactual, contardooptimal, forel2023explainable}, Discount \citep{you2024discount} \\ \hline
\textbf{Model} \(f\) & {\footnotesize Bagging, LightGBM, Support Vector Machine (SVM), Gaussian Process (GP), Radial Basis Function Network (RBF), XGBoost, Deep Neural Network (DNN), Random Forest (RndForest), AdaBoost, Gradient Boosting (GradBoost), Logistic Regression (LR), Quadratic Discriminant Analysis (QDA)} \\ \bottomrule
\end{tabular}
\label{tab:experiment-setup}%
\end{table}
\begin{table}[h!]
\centering
\caption{[\textit{Experiment Methods Setup}] The table defines $6$ methods for comparisons, colored to align with \figurename~\ref{fig:cola_results}. Each method is put in an experiment scenario, as defined in \tablename~\ref{tab:experiment-setup}, for benchmarking. }
\vspace{-2mm}
\resizebox{\textwidth}{!}{
\begin{tabular}{l|l|l}
\toprule
\textbf{Method}            & The probability $\vec{p}$ used by $\A_{\text{Value}}$ and  $A_{\text{Shap}}$                                       & The Shapley algorithm $\A_{\text{Shap}}$                                               \\ \hline
\textcolor{cyan}{\textbf{RB-${p}_{\text{Uni}}$}}   & $\vec{p}\leftarrow\A_{\text{Prob}}$:$p_{ij} = 1/nm$  ($\forall i,j$)                   & \small{RB-SHAP}, $\mathcal{D} =$ trainset of $f$                            \\ 
\textcolor{blue}{\textbf{RB-${p}_{\text{OT}}$}}    & $\vec{p}\leftarrow\A_{\text{Prob}}$:Eq.~\eqref{eq:ot} (but not used in $\A_{\text{Shap}}$)    & \small{RB-SHAP}, $\mathcal{D} =$ trainset of $f$     \\ 
\textcolor{orange}{\textbf{CF-$p_{\text{Uni}}$}}   & $\vec{p}\leftarrow\A_{\text{Prob}}$:$p_{ij} = 1/nm$  ($\forall i,j$)                  & \small{CF-SHAP}, $\mathcal{D}(\vec{x}_i) = \vec{p}_i$            \\
\textcolor{green}{\textbf{CF-${p}_{\text{Rnd}}$}}   & $\vec{p}\leftarrow\A_{\text{Prob}}$:Any $\vec{x}_i$ matched randomly to an $\vec{r}_j$               & \small{CF-SHAP}, $\mathcal{D}(\vec{x}_i) = \vec{p}_i$        \\ 
\textcolor{red}{\textbf{CF-${p}_{\text{OT}}$}}    & $\vec{p}\leftarrow\A_{\text{Prob}}$:Eq.~\eqref{eq:ot}                       & \small{\gls{p-shap}} with $\vec{p}$                                                       \\ 
\textbf{CF-$p_{\text{Ect}}$}   & $\vec{p}\leftarrow\A_{\text{Prob}}$:Any $\vec{x}_i$ matched to known counterpart $\vec{r}_j$        & \small{(CF or B)-SHAP}, $\D(\vec{x}_i)\rightarrow\vec{r}_j$    \\ \bottomrule
\end{tabular}
}
\label{tab:experiment-methods-setup}
\end{table}

 We briefly introduce the many $\A_{\text{CE}}$ in \tablename~\ref{tab:experiment-setup}. DiCE and KNN are data-instance-based \gls{ce} methods, which yield counterfactual(s) respectively for each factual instance. AReS and GlobeCE are group-based \gls{ce} methods, which find a collection of counterfactual instances for the whole factual data as a group. The algorithm Discount treats the factual instances as an empirical distribution and seeks a counterfactual distribution that stays in proximity to it.

\tablename~\ref{tab:experiment-methods-setup} defines 6 methods, where CF-$p_{\text{OT}}$ is the proposed $p$-SHAP and the others are baselines. Each is put in many experiment scenarios in \tablename~\ref{tab:experiment-setup}, benchmarked comprehensively. Each method is determined by a combination of the three algorithms $\A_{\text{Prob}}$, $\A_{\text{Value}}$, and $\A_{\text{Shap}}$. For example, the first row RB-$p_{\text{Uni}}$ uses uniform distribution as the algorithm $\A_{\text{Prob}}$ to compute $\vec{p}$, and then the computed $\vec{p}$ is sequentially used in \eqref{eq:A_value_max} or \eqref{eq:A_value_avg} of $\A_{\text{Value}}$ for computing $\vec{q}$, and $\A_{\text{Shap}}$ is RB-SHAP.

These methods are carefully designed for ablation studies. First, RB-$p_{\text{Uni}}$ differs with CF-$p_{\text{Uni}}$ in that the latter uses \gls{ce} information whereas the former does not. 
Second, CF-$p_{\text{Uni}}$, CF-$p_{\text{Rnd}}$, and CF-$p_{\text{OT}}$ use \gls{ce} in \gls{fa} with different joint distributions, and we demonstrate later that the distribution computed by \gls{ot} outperforms the others significantly. 
Third, we want to make sure that \gls{ot} is useful because it informs $\A_{\text{Shap}}$ with respect to the factual-counterfactual alignment, not because of other factors, and hence a comparison of CF-$p_{\text{OT}}$ that uses such an alignment to RB-$p_{\text{OT}}$ that does not. Finally, CF-$p_{\text{Ect}}$ represents a special case that each counterfactual originates from a known source, which is used as an exact factual-counterfactual alignment, making CF-SHAP also B-SHAP. 
\begin{table}[!h]
\centering
\caption{[\textit{\gls{cola} for Modification Minimality}] This table shows the number of modified features in $\vec{z}$ by each method, when $f(\vec{z})$ reaches 80\% counterfactual effect of $f(\vec{r})$. The result of each method is averaged by running in 4 randomly selected scenarios in \tablename~\ref{tab:experiment-setup}, with 100 runs in each scenario. The symbol ``-'' means the target counterfactual effect cannot be achieved.}
\vspace{-2mm}
\begin{tabular}{c|c|cc|cc}
\hline
\multirow{2}{*}{\textbf{Dataset}}                                                      & \multirow{2}{*}{\textbf{Method}} & \multicolumn{2}{c|}{\textbf{80\% Counterfactual Effect}}                                                    & \multicolumn{2}{c}{\textbf{100\% Counterfactual Effect}}                                            \\
                                                                                       &                                  & \textbf{\# Modified Features}                     & \textbf{$\frac{\|\vec{z}-\vec{x}\|}{\|\vec{r}-\vec{x}\|}$} & \textbf{\# Modified Features}             & \textbf{$\frac{\|\vec{z}-\vec{x}\|}{\|\vec{r}-\vec{x}\|}$} \\ \hline
\multirow{5}{*}{\begin{tabular}[c]{@{}c@{}}German \\ Credit \\ $|\F|=9$\end{tabular}}  & \textcolor{cyan}{\textbf{RB-${p}_{\text{Uni}}$}}                         & \multicolumn{1}{c|}{--}                         & --                                                          & \multicolumn{1}{c|}{--}                 & --                                                          \\
                                                                                       & \textcolor{blue}{\textbf{RB-${p}_{\text{OT}}$}}                           & \multicolumn{1}{c|}{$5.29(\pm 0.09)$}          & $75.9\%$                                                   & \multicolumn{1}{c|}{--}                 & --                                                          \\
                                                                                       & \textcolor{orange}{\textbf{CF-$p_{\text{Uni}}$}}                         & \multicolumn{1}{c|}{--}                         & --                                                          & \multicolumn{1}{c|}{--}                 & --                                                          \\
                                                                                       & \textcolor{green}{\textbf{CF-${p}_{\text{Rnd}}$}}                         & \multicolumn{1}{c|}{--}                         & --                                                          & \multicolumn{1}{c|}{--}                 & --                                                          \\
                                                                                       & \textcolor{red}{\textbf{CF-${p}_{\text{OT}}$}}                          & \multicolumn{1}{c|}{$\mathbf{1.70(\pm 0.02)}$} & $\mathbf{24.3\%}$                                          & \multicolumn{1}{c|}{$\mathbf{3.13(\pm 0.03)}$}  & $\mathbf{44.9\%}$                                                   \\ \hline
\multirow{5}{*}{\begin{tabular}[c]{@{}c@{}}Hotel \\ Bookings\\ $|\F|=29$\end{tabular}} & \textcolor{cyan}{\textbf{RB-${p}_{\text{Uni}}$}}                         & \multicolumn{1}{c|}{$7.10(\pm 0.08)$}          & $24.5\%$                                                   & \multicolumn{1}{c|}{--}                 & --                                                          \\
                                                                                       & \textcolor{blue}{\textbf{RB-${p}_{\text{OT}}$}}                           & \multicolumn{1}{c|}{$8.55(\pm 0.08)$}          & $50.2\%$                                                   & \multicolumn{1}{c|}{--}                 & --                                                          \\
                                                                                       & \textcolor{orange}{\textbf{CF-$p_{\text{Uni}}$}}                         & \multicolumn{1}{c|}{$7.01(\pm 0.07)$}          & $41.1\%$                                                   & \multicolumn{1}{c|}{--}                 & --                                                          \\
                                                                                       & \textcolor{green}{\textbf{CF-${p}_{\text{Rnd}}$}}                         & \multicolumn{1}{c|}{$10.63(\pm 0.08)$}         & $62.4\%$                                                   & \multicolumn{1}{c|}{--}                 & --                                                          \\
                                                                                       & \textcolor{red}{\textbf{CF-${p}_{\text{OT}}$}}                          & \multicolumn{1}{c|}{$\mathbf{2.50(\pm 0.03)}$} & $\mathbf{14.6\%}$                                          & \multicolumn{1}{c|}{$\mathbf{4.44(\pm 0.02)}$}  & $\mathbf{26.0\%}$                                                   \\ \hline
\multirow{5}{*}{\begin{tabular}[c]{@{}c@{}}COMPAS\\ $|\F|=15$\end{tabular}}            & \textcolor{cyan}{\textbf{RB-${p}_{\text{Uni}}$}}                         & \multicolumn{1}{c|}{$5.02(\pm 0.05)$}          & $82.7\%$                                                   & \multicolumn{1}{c|}{--}                 & --                                                          \\
                                                                                       & \textcolor{blue}{\textbf{RB-${p}_{\text{OT}}$}}                           & \multicolumn{1}{c|}{--}                         & --                                                          & \multicolumn{1}{c|}{--}                 & --                                                          \\
                                                                                       & \textcolor{orange}{\textbf{CF-$p_{\text{Uni}}$}}                         & \multicolumn{1}{c|}{$2.80(\pm 0.04)$}          & $34.4\%$                                                   & \multicolumn{1}{c|}{--}                 & --                                                          \\
                                                                                       & \textcolor{green}{\textbf{CF-${p}_{\text{Rnd}}$}}                         & \multicolumn{1}{c|}{$2.58(\pm 0.04)$}          & $32.1\%$                                                   & \multicolumn{1}{c|}{--}                 & --                                                          \\
                                                                                       & \textcolor{red}{\textbf{CF-${p}_{\text{OT}}$}}                          & \multicolumn{1}{c|}{$\mathbf{1.25(\pm 0.03)}$} & $\mathbf{14.8\%}$                                          & \multicolumn{1}{c|}{$\mathbf{2.45(\pm 0.03)}$}  & $\mathbf{30.0\%}$                                                   \\ \hline
\multirow{5}{*}{\begin{tabular}[c]{@{}c@{}}HELOC\\ $|\F|=23$\end{tabular}}             & \textcolor{cyan}{\textbf{RB-${p}_{\text{Uni}}$}}                         & \multicolumn{1}{c|}{--}                         & --                                                          & \multicolumn{1}{c|}{--}                 & --                                                          \\
                                                                                       & \textcolor{blue}{\textbf{RB-${p}_{\text{OT}}$}}                           & \multicolumn{1}{c|}{--}                         & --                                                          & \multicolumn{1}{c|}{--}                 & --                                                          \\
                                                                                       & \textcolor{orange}{\textbf{CF-$p_{\text{Uni}}$}}                         & \multicolumn{1}{c|}{--}                         & --                                                          & \multicolumn{1}{c|}{--}                 & --                                                          \\
                                                                                       & \textcolor{green}{\textbf{CF-${p}_{\text{Rnd}}$}}                         & \multicolumn{1}{c|}{$2.73(\pm 0.04)$}          & $15.7\%$                                                   & \multicolumn{1}{c|}{--}                 & --                                                          \\
                                                                                       & \textcolor{red}{\textbf{CF-${p}_{\text{OT}}$}}                          & \multicolumn{1}{c|}{$\mathbf{2.35(\pm 0.03)}$} & $\mathbf{13.4\%}$                                          & \multicolumn{1}{c|}{$\mathbf{7.745(\pm 0.05)}$} & $\mathbf{44.7\%}$                                                   \\ \hline
\end{tabular}
\label{tab:features}
\end{table}

\paragraph{Result I: \gls{cola} achieves significant action reduction with a minor loss in counterfactual effect} In \tablename~\ref{tab:features}, we set a goal for $\vec{z}$ such that $f(\vec{z})$ reaches 80\% or 100\% counterfacrtual effect of $f(\vec{r})$\footnote{Note that by definition, we have $D(f(\vec{r}),\vec{y}^*)=0$, which represents a 100\% counterfactual effect since $\vec{y}^*=f(\vec{r})$. To define a counterfactual effect 80\%, consider the proportion of divergence reduced by the refined \gls{ce} $\vec{z}$. That is,
\(
\textit{Counterfactual Effect} = 1 - {D(f(\vec{z}),y^*)}\big/{D(f(\vec{x}),y^*)} = 80\%
\), with $D$ being \gls{ot}.}. Observe that \gls{cola} is effective in achieving this goal, requiring significantly fewer actions in $\vec{z}$ (i.e., modifications of features in $\vec{z}$), compared to the original \gls{ce} $\vec{r}$. Using \gls{cola}, one could expect to only resort to 13\%--25\% of the feature changes (calculated by $\|\vec{z}-\vec{x}\|/\|\vec{r}-\vec{x}\|$) to achieve the counterfactual effect of 80\%. In particular, only \gls{cola} with \gls{p-shap} (that is, CF-$p_{\text{OT}}$) can reach the goal of the counterfactual effect of 100\%, with only 26\%--45\% of the feature changes in original $\vec{r}$. Especially, \gls{p-shap} leads to the best actional minimality, which is analyzed in details below.

%

\begin{figure}[t]
\begin{minipage}{\textwidth}
\centering
\includegraphics[width=1.01\linewidth]{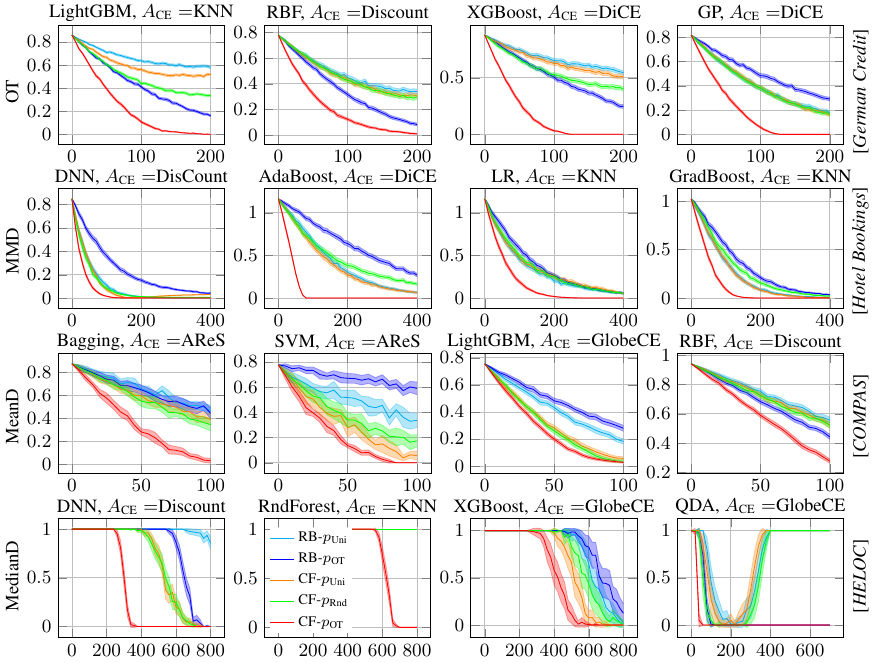}
\caption{$D(f(\vec{z}),\vec{y}^*)$ vs. allowed actions $C$. Experiments are with 100 runs. The shadows show the 99.9\% confidence intervals. $\A^{\text{avg}}_{\text{Value}}$ is used for \textit{HELOC} and \textit{COMPAS}, and $A^{\text{max}}_{\text{Value}}$ is used for \textit{German Credit} and \textit{Hotel Bookings}. The legend inside ``RnDForest, $A_{\text{CE}}=$KNN'' applies to all plots.
}\label{fig:cola_results}
\end{minipage}
\begin{minipage}{\textwidth}
\vspace{3mm}\centering
\includegraphics[width=1.01\linewidth]{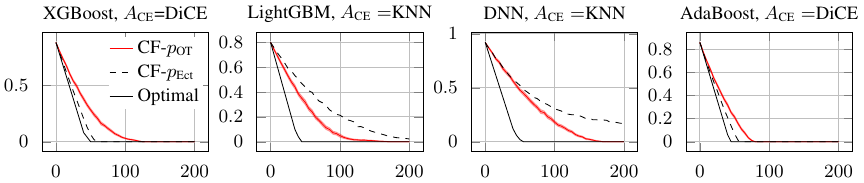}
\caption{[\textit{German Credit}] $D(f(z),y^*)$ vs. allowed actions $C$, with $D$ being \textit{\gls{meand}}.
}\label{fig:optimality}
\end{minipage}
\end{figure}

\paragraph{Result II: \gls{p-shap} outperforms the other Shapley methods in achieving counterfactual effect} We provide the evaluation of different Shapley methods in \eqref{eq:v_B}--\eqref{eq:v_f} in \figurename~\ref{fig:cola_results}, where the x-axis is the number of allowed feature changes $C$ and the y-axis is the term $D(f(\vec{z}), \vec{y}^*)$ in \eqref{eq:main_problem}. First, RB-$p_{\text{Uni}}$ and RB-$p_{\text{OT}}$ perform significantly worse than the others, indicating the importance of using \gls{ce} information in attributing features towards modification minimality in \gls{ce}. Second, the result showing that CF-$p_{\text{OT}}$ outperforms RB-$p_{\text{OT}}$ demonstrates that the use of \gls{ot} enhances the performance of \gls{p-shap} specifically by providing effective factual-counterfactual alignment, rather than being influenced by other factors, due to that they only differ in $\A_{\text{Shap}}$ (see \tablename~\ref{tab:experiment-methods-setup}). Third, \gls{p-shap} significantly outperforms CF-$p_{\text{Uni}}$ and CF-$p_{\text{Rnd}}$. This shows the effectiveness of the joint distribution obtained in \gls{ot} in using \gls{p-shap}. Namely, merely using the counterfactual information for \gls{fa} (as done by CF-$p_{\text{uni}}$ and CF-$p_{\text{Rnd}}$) is not enough, and a proper alignment (which does not necessarily mean the one defined by the exact counterfactual generation mechanism, as revealed in \figurename~\ref{fig:optimality} later) between factual and counterfactual must be considered.  

We observed that \gls{cola} with \gls{p-shap} possess good performance for many of the experiment scenarios defined in \tablename~\ref{tab:experiment-setup}, as shown in  \figurename~\ref{fig:cola_results} as well as \tablename~\ref{tab:features}. 
A massive amount of such results are further demonstrated in Appendix~\ref{subsec:extended_numerical}. In conclusion, \gls{p-shap} outperforms all the other Shapley methods for the sparsity in \gls{ce}.

\paragraph{Result III: \gls{cola} \textcolor{black}{is competitive with the \gls{milp} optimum on tractable benchmarks}} \textcolor{dtured}{This result demonstrates the effectiveness of $p$-SHAP in eliminating the influence of the \gls{ce} generation process by replacing the \gls{ce} algorithm-dependent knowledge\footnote{This knowledge, i.e. exact alignment between factual and counterfactual, is available only by DiCE and KNN among all \gls{ce} methods considered.} of $\mathcal{D}$ with the \gls{ot} joint distribution between the factual and counterfactual data, shown in \figurename~\ref{fig:optimality}.} We benchmark the method CF-$p_{\text{Ect}}$ using \gls{cola}, and focus on solving \eqref{eq:main_problem} with MeanD as the divergence function $D$. Since CF-$p_{\text{Ect}}$ relies on a known factual-counterfactual alignment, we benchmark the effectiveness of \gls{cola} for using this alginment. The theoretical optimality of \gls{cola} in this case can be obtained by solving an \gls{milp}, see Appendix~\ref{sec:ect_optimality} for how the \gls{milp} formulation is derived. Note that solving \gls{milp} is computationally heavy, hence only done for \textit{German Credit}. 

 We can see that CF-$p_{\text{Ect}}$ possess a near-optimal performance using DiCE. Remark that for DiCE and KNN, the factual-counterfactual pairs are independent to each others and hence we argue that \gls{cola} is effective for instance-based \gls{ce}, even though our formulation in \eqref{eq:main_problem} is a generalization for group or distributional \gls{ce}. It is interesting to note that CF-$p_{\text{OT}}$ sometimes performs better than CF-$p_{\text{Ect}}$, \textcolor{dtured}{because it utilizes a more theoretically grounded approach to identify the key features that require modification, whereas CF-$p_{\text{Ect}}$ relies on \gls{ce} algorithm-dependent knowledge, which lacks solid justification on its effectiveness for \gls{fa}.}
 We notice that there is still a gap between CF-$p_{\text{Ect}}$, CF-$p_{\text{OT}}$ and the optimal result in KNN. Finding the best alignment is still an open question.

\section{Conclusions}
\label{sec:conclusions}

This paper introduces a novel framework, \gls{cola}, for refining \gls{ce} by  joint-distribution-informed Shapley values, ensuring the refined \gls{ce} maintains the counterfactual effect with fewer actions. 

\section*{Ethics Statement}
All authors have read and agree to abide by the ICLR Code of Ethics.\footnote{\url{https://iclr.cc/public/CodeOfEthics}} This paper proposes a method (\emph{COLA} with \emph{p\mbox{-}SHAP}) for refining \gls{ce}. Our work is methodological and empirical; it does not involve human subjects or interventions, and we did not collect, annotate, or release any new personal data.

All experiments use standard, de-identified, publicly available tabular datasets (e.g., Adult/Census Income, German Credit, HELOC). We follow dataset licenses/terms and the common research practice of using these data solely for benchmarking. We do not attempt re-identification or linkage, and we release no additional personal information. Our code will include options to exclude protected attributes and to enforce feasibility/immutability constraints on features.

\gls{ce} can surface biases in underlying models but do not by themselves guarantee fairness. Our method can reduce the number of recommended feature edits; it does \emph{not} authorize edits to sensitive or immutable attributes (e.g., sex, race, age) nor does it confer causal validity. We discourage use of CE to “game” high-stakes systems (e.g., lending, hiring, healthcare) or to pressure individuals into unrealistic or unethical behavior changes. Any deployment in consequential domains should include domain-expert–defined feasibility constraints, fairness audits (e.g., disparate impact/benefit across groups), human oversight, and compliance with applicable data-protection and AI regulations. We explicitly caution that our attribution is distributional (not causal) and must not be interpreted as causal effect.

Risks include (i) misinterpretation of attributions as causal, (ii) recommending infeasible or harmful edits, and (iii) enabling strategic manipulation. We mitigate these by (a) stating non-causal scope and limitations in the paper, (b) supporting cost/feasibility masks and immutable features in the implementation, (c) reporting performance across subgroups where relevant, and (d) providing guidance for responsible use in documentation.

We provide an anonymized implementation, detailed settings, and seeds to facilitate replication; we report model types, hyperparameters, and compute budgets, and we avoid result cherry-picking by using fixed evaluation protocols. To our knowledge, there are no conflicts of interest or sensitive sponsorships related to this submission; any funding disclosures will be provided after the double-blind review.

\section*{Reproducibility Statement}
We take reproducibility seriously and provide pointers to all necessary components. The COLA framework and p\mbox{-}SHAP attribution are specified with pseudocode and notation in the main paper (Algorithm~1; Method and Preliminaries sections), with all theoretical assumptions and complete proofs deferred to the appendix (theorems and lemmas with line-by-line proofs). 

Implementation details—hyperparameters, optimization settings, environment specifications, and random seeds—are documented in Appendix~\ref{sec:reproducibility}, which also describes the datasets used and the evaluation protocol (including data splits and preprocessing). Baselines (models and CE generators) and coupling choices are enumerated in the experimental sections and ablations, with their configurations mirrored in the supplemental materials. A downloadable code repository 
\[
\text{{\small \faGithub ~ \url{https://github.com/youlei202/XAI-COLA}}}
\]
contains scripts to reproduce all results end-to-end, including a requirements/environment file and seed-controlled runs; we also report hardware and runtime in Appendix~\ref{sec:reproducibility} to contextualize compute requirements, and provide ablation/sensitivity studies and the small-scale MILP benchmark setup in the appendix to support independent verification.



%

\bibliography{iclr2026_conference}
\bibliographystyle{iclr2026_conference}

\newpage
\appendix
\appendixpage

{
\small
\setcounter{tocdepth}{1}
\startcontents[sections]
\printcontents[sections]{l}{1}{}
}

\section{Comparison to Existing Approaches}
\label{subsec:related_work}

The authors in \citep{albini2022counterfactual} proposed CF-SHAP, which uses counterfactual data points as the background distribution for Shapley. Yet, it assumes the counterfactual data distribution is defined conditionally on each single data instance, which implies that there is a known probabilistic alignment between every factual and every counterfactual instance. Similar assumptions are made in \citep{merrick2020explanation,kommiya2021towards}.
 In many scenarios where global explanations are expected, this assumption fails. We note that the setup in \citep{albini2022counterfactual} for counterfactual data distribution is a special case of ours. The authors in \citep{kwon2022weightedshap} proposed WeightedSHAP, adding weights to features rather than treating them as contributing equally. Our proposed method weights the contributions for data points and can be straightforwardly extended to consider weighting both rows and columns. Literature \citep{kommiya2021towards} establishes a framework for utilizing both \gls{fa} and \gls{ce} for explainability. Yet, the \gls{ce}-based \gls{fa} have the same assumption as \citep{albini2022counterfactual}, making it difficult to generalize to group \citep{ley2023globe,ley2022global} or distributional \gls{ce} \citep{you2024discount} cases. Another relevant work is \citep{sharma2022counterfactual}, a method that models counterfactuals using a system’s causal structure and time-series predictors to fairly decompose a single observed change in a system-level metric into contributions from individual inputs. More importantly, the aforementioned literature does not address the minimal actions \gls{ce} problem, which is the focus of our paper. 
 
 The problems formulated in \citep{kanamori2022counterfactual,karimi2021algorithmic} are quite close to the one investigated in this paper. In \citep{karimi2021algorithmic} the authors minimize the cost of performing actions with assumptiosn of known \gls{scm}, which is rarely known in practice. The authors in \citep{kanamori2022counterfactual} pointed out that it remains open whether existing \gls{ce} methods can be used for solving that problem. An-\gls{milp}-solvers based approach is proposed for linear classifiers, tree ensembles, and deep ReLU networks, built upon the works \citep{ustun2019actionable, kanamori2020dace, parmentier2021optimal}. However, solving \gls{milp} is costly, which makes it difficult to scale. 

\section{$\mathcal{NP}$-hardness of the Problem in \eqref{eq:main_problem}}
\label{subsec:np-hard}

Theorem~\ref{thm:np-hard} below states that one does not expect a scalable exact algorithm for solving it generally, unless $\mathcal{P}=\mathcal{NP}$, and the hardness lies not only on the non-linearity of $f$, but also its combinatorial nature. 

\begin{theorem}
Problem~\eqref{eq:main_problem} is generally $\mathcal{NP}$-hard for non-trivial divergences. More specifically, it is hard even when $d=1$ for linear models.
\label{thm:np-hard}
\end{theorem}
\begin{proof}
Consider the \textit{\gls{sr} with a Cardinality Constraint} problem, defined as follows. Given a matrix \( \vec{W} \in \mathbb{R}^{m \times n} \), a target vector \( \vec{y}^* \in \mathbb{R}^m \), and a sparsity level \( K \in \mathbb{N} \), the goal is to find a vector \( \vec{z} \in \mathbb{R}^n \) that minimizes the residual error while having at most \( K \) non-zero elements:
\begin{align*}
\min_{\vec{z}} \quad & \| \vec{W} \vec{z} - \vec{y}^* \|_2^2 \,,\\
\text{s.t.} \quad & \| \vec{z} \|_0 \leq K \,,
\end{align*}
where \( \| \vec{z} \|_0 \) denotes the number of non-zero elements in \( \vec{z} \). This problem is known to be \(\mathcal{NP}\)-hard due to the combinatorial nature of selecting the subset of variables to include in the model.

We will map this \gls{sr} problem to our problem~\eqref{eq:main_problem} with the following settings. Let the number of features be \( d = 1 \)  and the number of instances be \( n \) (the same as the dimension of the \gls{sr} problem). Let the factual data be \( \vec{x} = \vec{0} \in \mathbb{R}^n \) (the zero vector), and the target model output be \( \vec{y}^* \in \mathbb{R}^m \) (as given in the \gls{sr} problem). We define the model \( f \) as a linear function \( f(\vec{z}) = \vec{W} \vec{z} \).

For the model output, we define \( D(f(\vec{z}), \vec{y}^*) = \| f(\vec{z}) - \vec{y}^* \|_2^2 \), i.e. the Euclidean distance squared. For the instances, we define \( D(\vec{z}, \vec{x}) = \| \vec{z} - \vec{x} \|_2^2 = \| \vec{z} \|_2^2 \), since \( \vec{x} = \vec{0} \). We set the maximum allowed feature changes \( C \) to be equal to \( k \) (the sparsity level from the \gls{sr} problem). The large constant \( M \) can be any sufficiently large positive number, for example, \( M \geq \max_{i} | z_i | \).

Given that \( \vec{x} = \vec{0} \), constraints~\eqref{eq:main_problem_constr} and \eqref{eq:main_problem_constr_2} are simplified to: 
\begin{align*}
z_i &\leq 0 \cdot (1 - c_i) + M c_i = M c_i \,, \\
z_i &\geq 0 \cdot (1 - c_i) - M c_i = -M c_i \,, \quad \forall\, k = 1, \ldots, n \,.
\end{align*}
This means that if \( c_i = 0 \), then \( z_i \leq 0 \) and \( z_i \geq 0 \), so \( z_i = 0 \). If \( c_i = 1 \), then \( z_i \in [ -M, M ] \). The constraint~\eqref{eq:main_problem_C} becomes \( \sum_{k=1}^n c_i \leq k \).

Our problem~\eqref{eq:main_problem} thus becomes:
\begin{subequations}
\begin{align}
\min_{\vec{c}, \vec{z}} \quad & \| W \vec{z} - \vec{y}^* \|_2^2 \,,\label{eq:obj_revised} \\
\text{s.t.} \quad & \| \vec{z} \|_2^2 \leq \epsilon \,,\label{eq:epsilon_revised} \\
&\textstyle \sum_{k=1}^d c_i \leq k \,,\label{eq:C_revised} \\
& z_i = 0 \,, \quad\hspace{11.5mm} \text{if } c_i = 0 \,,\label{eq:z_zero_if_c_zero} \\
& z_i \in [ -M, M ] \,, \quad \text{if } c_i = 1 \,,\label{eq:z_range_if_c_one} \\
& c_i \in \{0, 1\} \,, \quad\hspace{5mm} \forall k = 1, \ldots, n \,. \label{eq:c_binary}
\end{align}
\label{eq:np-hard-revised}
\end{subequations}

In this formulation, the variables \( c_i \) indicate whether the variable \( z_i \) is allowed to change (\( c_i = 1 \)) or not (\( c_i = 0 \)). The constraints enforce that \( z_i = 0 \) when \( c_i = 0 \), mirroring the sparsity constraint in the \gls{sr} problem. The constraint \( \sum_{k=1}^n c_i \leq k \) ensures that at most \( k \) features can change, matching the sparsity level. The objective function is identical to that of \gls{sr}.

Since our problem formulation directly mirrors the \gls{sr} problem with a cardinality constraint, which is known to be \(\mathcal{NP}\)-hard, solving Problem~\eqref{eq:main_problem} is at least as hard as solving the \gls{sr} problem. Therefore, Problem~\eqref{eq:main_problem} is \(\mathcal{NP}\)-hard even when \( d = 1 \), the model \( f \) is linear, and the divergence functions \( D \) are standard Euclidean distances.
\end{proof}

\section{Proof of Theorem~\ref{thm:lipschitz}: \gls{p-shap} Towards Counterfactual Effect}
\label{subsec:ot_theory}

 Let \( \vec{x} = \{ \vec{x}_i \}_{i=1}^n \in \mathbb{R}^{n\times d} \) be the set of factual data points with associated probability weights \( \mu_i \geq 0 \) such that \( \sum_{i=1}^n \mu_i = 1 \), and let \( \vec{r} = \{ \vec{r}_j \}_{j=1}^m \in \mathbb{R}^{m\times d} \) be the set of counterfactual data points with associated probability weights \( \nu_j \geq 0 \) such that \( \sum_{j=1}^m \nu_j = 1 \). Let \( \vec{p}_{\text{OT}} \in \mathbb{R}^{n \times m} \) be the \gls{ot} plan between \( \vec{x} \) and \( \vec{r} \) that minimizes the expected transportation cost:
\[
\vec{p}_{\text{OT}} = \arg\min_{\vec{p} \in \Pi(\bm{\mu}, \bm{\nu})} \sum_{i=1}^n \sum_{j=1}^m p_{ij} \| \vec{x}_i - \vec{r}_j \|_2^2 \,,
\]
where \( \Pi(\bm{\mu}, \bm{\nu}) \) is the set of joint distributions satisfying the marginal constraints \( \sum_{j=1}^m p_{ij} = \mu_i \) for all \( i \) and \( \sum_{i=1}^n p_{ij} = \nu_j \) for all \( j \).
The theorem below provides that feature attributions are aligned with the expected costs of feature modifications, leading to action plans that are cost-efficient in achieving counterfactual outcomes.

\begin{theorem}[Theorem~\ref{thm:lipschitz} in the main text]
Consider the $1$-Wasserstein divergence $W_1$, i.e.
\(
W_1(f(\vec{x}),\vec{y}^*)=\min_{\bm{\pi}\in\Pi}\sum_{i=1}^n \sum_{j=1}^m \pi_{ij} \left| f(\vec{x}_i) - \vec{y}^*_j \right|
\).
Suppose the counterfactual outcome $\vec{y}^*$ is fully achieved by $\vec{r}$, i.e. $\vec{y}^*_j = f(\vec{r}_j)$ ($j=1,2\ldots, m$).
Assume that the model \( f: \mathbb{R}^d \rightarrow \mathbb{R} \) is Lipschitz continuous with Lipschitz constant \( L \).
The expected absolute difference in model outputs between the factual and counterfactual instances, weighted by \( \vec{p}_{\text{OT}} \) (with $\varepsilon=0$), is bounded by:
\[
W_1(f(\vec{x}),\vec{y}^*) \leq L \sqrt{ \sum_{i=1}^n \sum_{j=1}^m p_{ij}^{\text{OT}} \| \vec{x}_i - \vec{r}_j \|_2^2 }\leq L \sqrt{ \sum_{i=1}^n \sum_{j=1}^m p_{ij} \| \vec{x}_i - \vec{r}_j \|_2^2 } \quad \forall \vec{p}\in\Pi \,.
\]
Namely, $\vec{p}_{\text{OT}}$ minimizes the upper bound of $W_1(f(\vec{x}),\vec{y}^*)$, where the upper bound is based on the expected feature modification cost.
\label{thm:lipschitz-appendix}
\end{theorem}
\begin{proof}
Since the model \( f \) is Lipschitz continuous with constant \( L \), for any \( \vec{x}_i \in \mathbb{R}^d \) and \( \vec{r}_j \in \mathbb{R}^d \), it holds that:
\[
| f(\vec{x}_i) - f(\vec{r}_j) | \leq L \| \vec{x}_i - \vec{r}_j \|_2 \,.
\]
Multiplying both sides of the inequality by \( p_{ij}^{\text{OT}} \geq 0 \), we obtain:
\[
p_{ij}^{\text{OT}} | f(\vec{x}_i) - f(\vec{r}_j) | \leq L p_{ij}^{\text{OT}} \| \vec{x}_i - \vec{r}_j \|_2 \,.
\]
Summing both sides over all \( i = 1, \dots, n \) and \( j = 1, \dots, m \), we have:
\[
\sum_{i=1}^n \sum_{j=1}^m p_{ij}^{\text{OT}} | f(\vec{x}_i) - f(\vec{r}_j) | \leq L \sum_{i=1}^n \sum_{j=1}^m p_{ij}^{\text{OT}} \| \vec{x}_i - \vec{r}_j \|_2 \,.
\]
Let us denote \( E_f = \sum_{i=1}^n \sum_{j=1}^m p_{ij}^{\text{OT}} | f(\vec{x}_i) - f(\vec{r}_j) | \) and \( E_d = \sum_{i=1}^n \sum_{j=1}^m p_{ij}^{\text{OT}} \| \vec{x}_i - \vec{r}_j \|_2 \). The inequality then becomes:
\[
E_f \leq L E_d \,.
\]
To further bound \( E_d \), we apply the Cauchy-Schwarz inequality. Observe that the weights \( p_{ij}^{\text{OT}} \) are non-negative and satisfy \( \sum_{i=1}^n \sum_{j=1}^m p_{ij}^{\text{OT}} = 1 \) because \( p_{\text{OT}} \) is a probability distribution over the joint space of \( \vec{x} \) and \( \vec{r} \). The Cauchy-Schwarz inequality states that for any real-valued functions \( a_{ij} \) and \( b_{ij} \),
\[
\left( \sum_{i,j} a_{ij} b_{ij} \right)^2 \leq \left( \sum_{i,j} a_{ij}^2 \right) \left( \sum_{i,j} b_{ij}^2 \right).
\]
Setting \( a_{ij} = \sqrt{p_{ij}^{\text{OT}}} \) and \( b_{ij} = \sqrt{p_{ij}^{\text{OT}}} \| \vec{x}_i - \vec{r}_j \|_2 \), we have:
\[
E_d = \sum_{i,j} p_{ij}^{\text{OT}} \| \vec{x}_i - \vec{r}_j \|_2 = \sum_{i,j} \sqrt{p_{ij}^{\text{OT}}} \sqrt{p_{ij}^{\text{OT}}} \| \vec{x}_i - \vec{r}_j \|_2 = \sum_{i,j} a_{ij} b_{ij} \,.
\]
Applying the Cauchy-Schwarz inequality:
\[
E_d^2 \leq \left( \sum_{i,j} a_{ij}^2 \right) \left( \sum_{i,j} b_{ij}^2 \right) = \left( \sum_{i,j} p_{ij}^{\text{OT}} \right) \left( \sum_{i,j} p_{ij}^{\text{OT}} \| \vec{x}_i - \vec{r}_j \|_2^2 \right).
\]
Since \( \sum_{i,j} p_{ij}^{\text{OT}} = 1 \), this simplifies to:
\[
E_d^2 \leq \sum_{i=1}^n \sum_{j=1}^m p_{ij}^{\text{OT}} \| \vec{x}_i - \vec{r}_j \|_2^2 = \sum_{i=1}^n \sum_{j=1}^m p_{ij}^{\text{OT}} \| \vec{x}_i - \vec{r}_j \|_2^2 \,.
\]
Taking the square root of both sides, we obtain:
\[
E_d \leq \sqrt{ \sum_{i=1}^n \sum_{j=1}^m p_{ij}^{\text{OT}} \| \vec{x}_i - \vec{r}_j \|_2^2 } \,.
\]
Substituting back into the inequality for \( E_f \), we have:
\[
E_f \leq L \sqrt{ \sum_{i=1}^n \sum_{j=1}^m p_{ij}^{\text{OT}} \| \vec{x}_i - \vec{r}_j \|_2^2 } \,.
\]
Note that the $1$-Wasserstein divergence is no more than $E_f$\footnote{Note that \( E_f = \sum_{i=1}^n \sum_{j=1}^m p_{ij}^{\text{OT}} | f(\vec{x}_i) - f(\vec{r}_j) | \). Because both $\bm{\pi}$ and $\vec{p}_{\text{OT}}$ denote joint probability of $\vec{x}$ and $\vec{r}$, however, $\bm{\pi}$ makes the summation the minimum for the $W_1$ term across all possible joint distributions, whereas $\vec{p}_{\text{OT}}$ does not.}, then, 
\[
W_1(f(\vec{x}), \vec{y}^*)=\min_{\bm{\pi}\in\Pi}\sum_{i=1}^n \sum_{j=1}^m \pi_{ij} \left| f(\vec{x}_i) - \vec{y}^*_j \right|\leq E_f \leq L \sqrt{ \sum_{i=1}^n \sum_{j=1}^m p_{ij}^{\text{OT}} \| \vec{x}_i - \vec{r}_j \|_2^2 } \,.
\]
Therefore, the $1$-Wasserstein divergence between \( f(\vec{x}) \) and \( \vec{y}^* \) is bounded by the Lipschitz constant \( L \) times the square root of the expected transportation cost under \( \vec{p}_{\text{OT}} \).
Since \( \vec{p}_{\text{OT}} \) minimizes the expected transportation cost \( \sum_{i,j} p_{ij} \| \vec{x}_i - \vec{r}_j \|_2^2 \) over all feasible transport plans in \( \Pi(\bm{\mu}, \bm{\nu}) \), we have
\[
\sqrt{ \sum_{i=1}^n \sum_{j=1}^m p_{ij}^{\text{OT}} \| \vec{x}_i - \vec{r}_j \|_2^2 }\leq  \sqrt{ \sum_{i=1}^n \sum_{j=1}^m p_{ij} \| \vec{x}_i - \vec{r}_j \|_2^2 } 
\,,\quad \forall \vec{p}\in\Pi(\bm{\mu},\bm{\nu}) \,.
\]
Hence the conclusion follows.
\end{proof}

\section{Proof of Theorem~\ref{thm:intervention}: Interventional Effect}
\label{sec:proof_thm_intervention}

\begin{theorem}[Theorem~\ref{thm:intervention} in the main text]
For any subset \( S \subseteq \F \) and any \( \vec{x}_i \) (\(i=1,2,\ldots,n\)), \( v^{(i)}(S) \) represents the causal effect of the difference between the expected value of \( f(\vec{r}) \) under the intervention on features \( S \) and the unconditional expected value of \( f(\vec{r}) \). Mathematically, this is expressed as:
\[
\EE[f(\vec{r})]+v^{(i)}(S) =  \EE\left[f(\vec{r}) \middle| do\left(\vec{r}_S = \vec{x}_{i,S}\right) \right].
\]
\end{theorem}

\begin{proof}
Let $p(\vec{x},\vec{r})$ be the joint probability of $\vec{x}$ and $\vec{r}$ obtained from $\A_{\text{Prob}}$.
Under the intervention $do(\vec{r}_S=\vec{x}_{i,S})$, the features in $S$ are set to $\vec{x}_{i,S}$, and the features in $\F\backslash S$ remain distributed according to their marginal distribution $p(\vec{r}_{\F\backslash S})$. Therefore, the expected value of $f(\vec{r})$ under the intervention is:
\[
\EE[f(\vec{r}) \mid do(\vec{r}_S = \vec{x}_{i,S})] = \int_{\mathcal{R}_{\mathcal{F} \backslash S}} f(\vec{x}_{i,S};\vec{r}_{\F\backslash S}) p(\vec{r}_{\mathcal{F} \backslash S}) \, d \vec{r}_{\mathcal{F} \backslash S} \,.
\]
Remark that by the definition of $v^{(i)}(S)$,
\[
  v^{(i)}(S) = \EE_{\vec{r}\sim p(\vec{r}|\vec{x}_i)}[f(\vec{x}_{i,S};\vec{r}_{\F\backslash S})] - \EE[f(\vec{r})] = \int_{\mathcal{R}_{\mathcal{F} \backslash S}} f(\vec{x}_{i,S};\vec{r}_{\F\backslash S}) p(\vec{r}_{\mathcal{F} \backslash S}) \, d \vec{r}_{\mathcal{F} \backslash S}-\EE[f(\vec{r})],
\]
such that
\[
\EE[f(\vec{r})]+v^{(i)}(S) =  \EE \left[f(\vec{r}) \mid do\left(\vec{r}_S = \vec{x}_{i,S}\right) \right].
\]
Hence the conclusion.
\end{proof}
The value function \( v^{(i)}(S) \) captures the expected value of \( f(\vec{r}) \) when we intervene by setting the features in \( S \) to \(\vec{x}_{i,S}\), denoted as \( do(\vec{r}_S = \vec{x}_{i,S}) \). This intervention is independent of any predefined joint probability distribution $p(\vec{x},\vec{r})$. Therefore, the expression \( \mathbb{E}[f(\vec{r})] + v^{(i)}(S) \) represents the combined effect of the base expected value of \( f(\vec{r}) \) and the additional causal impact of the attribution \( v^{(i)}(S) \).

\section{Proof of Theorem~\ref{thm:z_distance}: Counterfactual Proximity}
\label{sec:z_distance_proof}

\begin{theorem}[Theorem~\ref{thm:z_distance} in the main text]
\textcolor{black}{Let \( \vec{q} \in \mathbb{R}^{n \times d} \) be the aligned reference matrix derived from the counterfactual set \( \vec{r} \in \mathbb{R}^{m \times d} \) via the value assignment step \( \vec{q} = A_{\text{Value}}(\vec{r}, \vec{p}) \), where \( \vec{p} \) is the coupling matrix. For any dimensions \( n, m \ge 1 \), the refined counterfactual \( \vec{z} \) constructed by the COLA framework satisfies:
\[
\| \vec{z} - \vec{x} \|_F \leq \| \vec{q} - \vec{x} \|_F \,.
\]
This inequality guarantees that the refined counterfactual \( \vec{z} \) is strictly closer (or equal) to the factual input \( \vec{x} \) compared to the aligned counterfactual proposal \( \vec{q} \), ensuring that the refinement process introduces no additional divergence.
}
\end{theorem}
\begin{proof}
\textcolor{black}{Let \( \vec{q} \) be the aligned reference matrix generated by \( \A_{\text{Value}}(\vec{r}, \vec{p}) \).}
For the elements where \( c_{ik} = 1 \) (modified elements), \( z_{ik} = \textcolor{black}{q_{ik}} \). For elements where \( c_{ik} = 0 \), \( z_{ik} = x_{ik} \). Therefore, we can write:
\begin{equation}
(z_{ik} - x_{ik})^2 = \begin{cases}
\textcolor{black}{(q_{ik} - x_{ik})^2} \,, & \text{if } c_{ik} = 1 \,; \\
0 \,, & \text{if } c_{ik} = 0 \,.
\end{cases}
\label{eq:z-x-r-x}
\end{equation}

The squared Frobenius norms with respect to \textcolor{black}{\(\vec{q}\)} and \(\vec{z}\) are computed as follows:
\[ \begin{split}
{ \color{black}
\| \vec{q} - \vec{x} \|_F^2 = \sum_{i=1}^n \sum_{k=1}^d (q_{ik} - x_{ik})^2 
} \,,\\
\| \vec{z} - \vec{x} \|_F^2 = \sum_{i=1}^n \sum_{k=1}^d (z_{ik} - x_{ik})^2 \,.
\end{split} \]
And,
\begin{align*}
\textcolor{black}{\| \vec{q} - \vec{x} \|_F^2} - \| \vec{z} - \vec{x} \|_F^2 &= \textcolor{black}{\sum_{i=1}^n \sum_{k=1}^d (q_{ik} - x_{ik})^2} - \sum_{i=1}^n \sum_{k=1}^d (z_{ik} - x_{ik})^2 \\
&\overset{(i)}{=} \textcolor{black}{\sum_{i=1}^n \sum_{k=1}^d (q_{ik} - x_{ik})^2} - \textcolor{black}{\sum_{(i,k): c_{ik} = 1} (q_{ik} - x_{ik})^2} \\
&= \textcolor{black}{\sum_{(i,k): c_{ik} = 0} (q_{ik} - x_{ik})^2} \geq 0 \,,
\end{align*}
where the equality (i) holds because of \eqref{eq:z-x-r-x}.

Since the difference \( \textcolor{black}{\| \vec{q} - \vec{x} \|_F^2} - \| \vec{z} - \vec{x} \|_F^2 \geq 0 \), it follows that:
\[
\| \vec{z} - \vec{x} \|_F^2 \leq \textcolor{black}{\| \vec{q} - \vec{x} \|_F^2} \,.
\]
Taking square roots, we get:
\[
\| \vec{z} - \vec{x} \|_F \leq \textcolor{black}{\| \vec{q} - \vec{x} \|_F} \,.
\]
Hence the conclusion.
\end{proof}
\begin{corollary}[Corollary~\ref{cor:z_distance_permutation} in the main text]
\textcolor{black}{
In the special case where \( n = m \) and the OT plan \( \vec{p} \) corresponds to a deterministic permutation \( \sigma \) (i.e., obtained without entropic regularization, \( \varepsilon = 0 \)), the matrix \( \vec{q} \) becomes a row-permuted version of \( \vec{r} \) (denoted as \( \vec{r}_{\sigma} \), where \( q_{i} = r_{\sigma(i)} \)). In this setting, the bound simplifies to:
\[
\| \vec{z} - \vec{x} \|_F \leq \| \vec{r}_{\sigma} - \vec{x} \|_F \,,
\]
recovering the intuition that the refined counterfactual is at least as close to the factuals as the original counterfactual set reordered by \( \sigma \).
}
\end{corollary}
\begin{proof}
\textcolor{black}{
Under the assumptions \( n=m \) and \( \varepsilon=0 \), the optimal transport plan \( \vec{p} \) becomes a permutation matrix associated with a bijection \( \sigma \). Consequently, the value assignment algorithm \( A_{\text{Value}} \) (whether maximizing or averaging) assigns \( \vec{q}_i = \vec{r}_{\sigma(i)} \) for every instance \( i \). Substituting \( \vec{q} \) with the permuted matrix \( \vec{r}_{\sigma} \) into the general bound established in Theorem~\ref{thm:z_distance} (i.e., \( \| \vec{z} - \vec{x} \|_F \leq \| \vec{q} - \vec{x} \|_F \)) directly yields the result.}
\end{proof}

\section{Analysis of Computational Complexity of \gls{cola}}
\label{subsec:complexity}

We perform analysis of the computational complexity of \gls{cola} as follows.

  First, we analyze $\A_{\text{Prob}}$. If the alignment between $\vec{x}$ and $\vec{r}$ is known a priori, then $\A_{\text{Prob}}$ just constructs the matrix $\vec{p}$ with the prior knowledge, which takes $\mathcal{O}(n\times m)$. Otherwise, we consider solving \gls{ot} to obtain $\vec{p}$, which, by the Sinkhorn–Knopp algorithm, takes $\mathcal{O}(n\times m\times \log(1/\varepsilon))$. 

 Then we analyze $\A_{\text{Shap}}$. For each subset, the model is evaluated on all $n$ data points, leading to $\mathcal{O}(n)$ evluations per subset. Incorporating baseline values from the reference data $\vec{r}$ involves replacing the values of certain features with their corresponding baseline values. This operation is $\mathcal{O}(m)$ because it requires accessing the baseline values from the reference table $\vec{r}$ for each of the $d$ features. If we assume that the reference values can be precomputed and accessed in constant time, then the complexity of incorporating these values can be considered as $\mathcal{O}(d)$. The number of $M_{\text{Shap}}$ subsets results in $M_{\text{Shap}}$ model evaluations. Combining the above steps, the complexity of $\A_{\text{Shap}}$ is $\mathcal{O}(n\times d\times M_{\text{Shap}})$. 

The normalization in line~\ref{alkg:cola-phi} of \gls{cola} takes $\mathcal{O}(n\times d)$\,.

To compare the complexities of the two algorithms, $\A^{\text{max}}_{\text{Value}}$ and $\A^{\text{avg}}_{\text{Value}}$, we analyze each algorithm step-by-step. For $\A^{\text{max}}_{\text{Value}}$, for each row $\vec{x}_i$ in the data table, we need to (1) compute the probabilities $p_{ij}$ for all $j \in \{1, 2, \ldots, m\}$, which involves $\mathcal{O}(m)$ operations per row, (2) identify the row $\vec{r}_j$ in the reference data with the highest probability $p_{ij}$, which involves $\mathcal{O}(m)$ operations per row, and (3) assign $q_{ik} = r_{\tau(i), k}$ where $\tau(i) = \arg\max_{j} p_{ij}$, which involves $\mathcal{O}(d)$ operations per row. Since there are $n$ rows in the data table, the total complexity for $\A^{\text{max}}_{\text{Value}}$ is $\mathcal{O}(n \times (m + m + d)) = \mathcal{O}(n \times (2m + d)) = \mathcal{O}(n \times m + n \times d) = \mathcal{O}(n \times (m + d))$.

For $\A^{\text{avg}}_{\text{Value}}$, for each row $\vec{x}_i$ in the data table, we need to 1) compute the probabilities $p_{ij}$ for all $j \in \{1, 2, \ldots, m\}$, which involves $\mathcal{O}(m)$ operations per row, 2) compute the sum $\sum_{j'=1}^m p_{ij'}$, which involves $\mathcal{O}(m)$ operations per row, and 3) calculate the weighted average $q_{ik} = \sum_{j=1}^m \left(\frac{p_{ij}}{\sum_{j'=1}^m p_{ij'}}\right) r_{jk}$, which involves $\mathcal{O}(m \times d)$ operations per row. Since there are $n$ rows in the data table, the total complexity for $\A^{\text{avg}}_{\text{Value}}$ is $\mathcal{O}(n \times (m + m + m \times d)) = \mathcal{O}(n \times (2m + m \times d)) = \mathcal{O}(n \times (m + m \times d)) = \mathcal{O}(n \times m \times (1 + d)) = \mathcal{O}(n \times m \times d)$.

For lines~\ref{alg:cola-c-begin}--\ref{alg:cola-c-end}, the entire complexity is straightforwardly $\mathcal{O}(n\times d) + \mathcal{O}(C)=\mathcal{O}(n\times d)$ due to the fact $C\leq n\times d$.

Therefore, the complexity of \gls{cola} using $\A^{\text{max}}_{\text{Value}}$ equals
\begin{align*}
    & \mathcal{O}(M_{\text{CE}}) + \mathcal{O}(n m \log(1/\varepsilon)) + \mathcal{O}(n d M_{\text{Shap}}) + \mathcal{O}(n d) + \mathcal{O}(n  (m + d)) + \mathcal{O}(n d) \\
=   & \mathcal{O}(M_{\text{CE}}) + \mathcal{O}(n m \log(1/\varepsilon)) + \mathcal{O}(n d M_{\text{Shap}}) + \mathcal{O}(nm) + \mathcal{O}(nd) \,,
\end{align*}
and the complexity of \gls{cola} using $\A^{\text{avg}}_{\text{Value}}$ equals 
\begin{align*}
    & \mathcal{O}(M_{\text{CE}}) + \mathcal{O}(n m \log(1/\varepsilon)) + \mathcal{O}(n d M_{\text{Shap}}) + \mathcal{O}(n d) + \mathcal{O}(nmd)) + \mathcal{O}(n d) \\
=   & \mathcal{O}(M_{\text{CE}}) + \mathcal{O}(n m \log(1/\varepsilon)) + \mathcal{O}(n d M_{\text{Shap}}) + \mathcal{O}(nmd) \,.
\end{align*}
Hence the complexity of \gls{cola} with respect to $n$, $m$, $d$, and the regularization parameter $\varepsilon$ of entropic \gls{ot} is 
\[
\mathcal{O}(M_{\text{CE}}) + \mathcal{O}(n m \log(1/\varepsilon)) + \mathcal{O}(n d M_{\text{Shap}}) + N \,,
\]
where $N=\mathcal{O}(nm) + \mathcal{O}(nd)$ if $\A^{\text{max}}_{\text{Value}}$ is used and $N=\mathcal{O}(nmd)$ if $\A^{\text{avg}}_{\text{Value}}$ is used.

\section{An \gls{milp} Formulation of \eqref{eq:main_problem} With \gls{meand}}
\label{sec:ect_optimality}

In this section, we provide a global optimality benchmark for using a known alignment between factual and counterfactual in solving \eqref{eq:main_problem} with $D$ being \gls{meand}, namely
\[
D(f(\vec{z}), \vec{y}^*) = \left|\frac{1}{n}\sum_{i=1}^n f(\vec{z}_i) - \widebar{y^*} \right| \,,
\]
with $\widebar{y^*}=\frac{1}{m}\sum_{j=1}^m y^*_j$.
Since \gls{cola} is used, we have $D(\vec{r},\vec{x})\leq \epsilon$, and $\vec{z}$ stays closer to $\vec{x}$ than $\vec{r}$, hence \eqref{eq:main_problem_epsilon} is dropped.
The formulation of \eqref{eq:main_problem} then becomes:
\begin{subequations}
\begin{align} 
\min_{\vec{c},\vec{z}}  \quad &  \left|\sum_{i=1}^n f(\vec{z}_i) - n \widebar{y^*} \right| \,,\\ 
\text{s.t.} \quad  & \sum_{i=1}^n\sum_{k=1}^d c_{ik} \leq C \,,\label{eq:ect_problem-C}\\
                   & z_{ik} = r_{ik}c_{ik}+ x_{ik} (1-c_{ik}) \,,  \quad i=1,\ldots, n \,,~\ k=1,\ldots, d \,.\label{eq:ect_problem-z}
\end{align}
\label{eq:ect_problem}%
\end{subequations}
Note that the original constraints~\eqref{eq:main_problem_constr} and~\eqref{eq:main_problem_constr_2} merge to be \eqref{eq:ect_problem-z}, because CF-$p_{\text{Ect}}$ is imposed to be used. That is, for any $\vec{x}_i$, there is an exact $\vec{r}_j$ serves as its reference in $\A_{\text{Shap}}$ and $\A_{\text{Value}}$, such that $x_{ik}$ ($k=1,2\ldots, d$) either stays unchanged or can be changed to $r_{jk}$. Therefore $z_{ik}=r_{ik}c_{ik}+ x_{ik} (1-c_{ik})$ of which the value depends on the binary variable $c_{ik}$.

Due to the known alignment between any $\vec{x}_i$ and its corresponding $\vec{r}_j$, $\vec{q}$ is determined (also, both $\A^{\text{max}}_{\text{Values}}$ and $\A^{\text{avg}}_{\text{Values}}$ return the same $\vec{q}$).
For any data point $i$ and any feature set $S\subseteq \F$, let $\vec{z}_{iS}$ denotes the solution $\vec{z}_i$ where we have all features $k\in S$ changed to $q_{ik}$, and the other features $h\in\F\backslash S$ stays $x_{ih}$. Hence the set of $\vec{z}_{iS}$ ($S\subseteq \F$) composes the domain of all possible values of $\vec{z}$. Define a corresponding scaler variable for any $\vec{z}_{iS}$:
\[
g_{iS} = f(\vec{z}_{iS}) - \widebar{y^*} \,. 
\]
Then, for any $\vec{z}_i$ in \eqref{eq:ect_problem}, the value of the term $\sum_{i=1}^n f(z_i) - n \widebar{y^*}$ can be represented by a binary variable $a_{iS}$ together with the scaler $g_{iS}$, namely,
\[
\sum_{i=1}^n f(\vec{z}_i) - n \widebar{y^*} = \sum_{i=1}^n \sum_{S\subseteq\F} g_{iS}a_{iS} \,.
\]
The optimization problem~\eqref{eq:ect_problem} is hence reformulated as a mixed integer programming below.
\begin{subequations}
\begin{align} 
\min_{\vec{a},\eta} \quad & \eta  \,,\\ 
\text{s.t.} \quad & \sum_{i=1}^n \sum_{S\subseteq\F} g_{iS}a_{iS} \leq \eta \,,\label{eq:optimality_baseline-eta}\\
                  & \sum_{i=1}^n \sum_{S\subseteq\F} g_{iS}a_{iS} \geq -\eta \,,\label{eq:optimality_baseline-neg-eta} \\
                  & \sum_{S\subseteq\F} a_{iS}= 1 \,, \quad i=1,\ldots, n \,,\label{eq:optimality_baseline-a}\\
                  & \sum_{i=1}^n\sum_{S\subseteq\F} |S|a_{iS} \leq C \,.\label{eq:optimality_baseline-C}
\end{align}
\label{eq:optimality_baseline}%
\end{subequations}
Minimizing $\eta$ under the two constraints \eqref{eq:optimality_baseline-eta} and \eqref{eq:optimality_baseline-neg-eta} is equivalent to minimizing the objective function of \eqref{eq:ect_problem}. The constraints in \eqref{eq:optimality_baseline-a} guarantees that each data point $i$ is subject to one and only one feature modification plan $a_{iS}$ for a specific $S$ ($S\subseteq \F$). The constraint \eqref{eq:optimality_baseline-C} corresponds to \eqref{eq:ect_problem-C}.
Solving \eqref{eq:optimality_baseline} yields the theoretical optimality of \gls{cola} using a known alignment between factual and counterfactual, demonstrated in \figurename~\ref{fig:optimality} in Section \ref{sec:numerical}.

\section{Extended Numerical Results}
\label{subsec:extended_numerical}

\begin{figure}[t]
\begin{minipage}{\linewidth}
\vspace{-2mm}\centering
\includegraphics[width=1.01\linewidth]{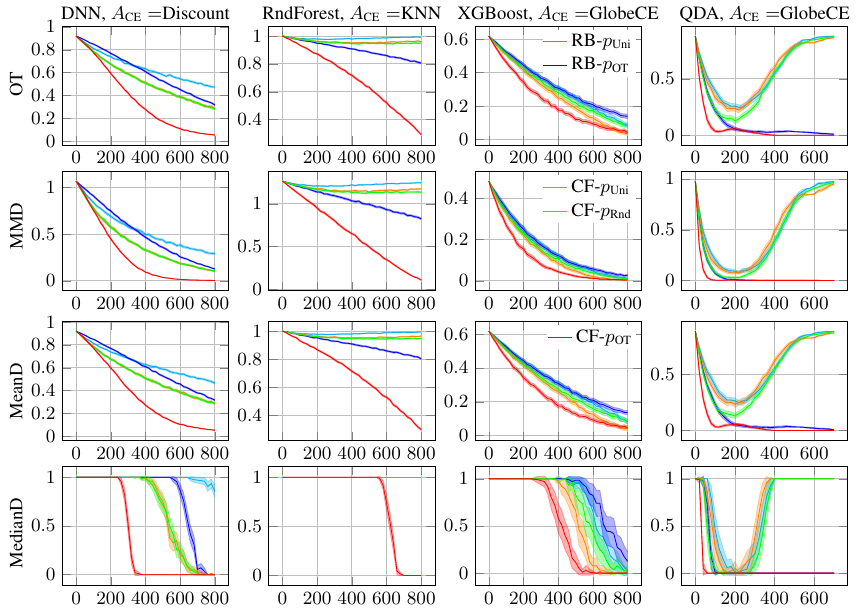}
\caption{[\textit{HELOC}] $D(f(\vec{z}),\vec{y}^*)$ vs. allowed actions $C$. Experiments are with 100 runs. The shadows show the 99.9\% confidence intervals. The legends apply to all plots. $\A^{\text{avg}}_{\text{Value}}$ is used.
}\label{fig:heloc}
\end{minipage}
\begin{minipage}{\linewidth}
\vspace{3mm}\centering
\includegraphics[width=1.01\linewidth]{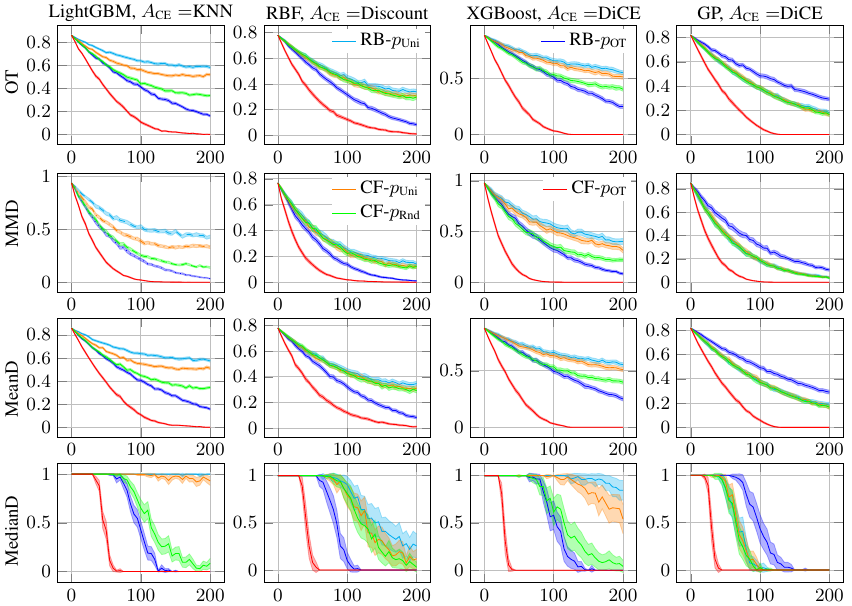}
\caption{[\textit{German Credit}] $D(f(\vec{z}),\vec{y}^*)$ vs. allowed actions $C$. Experiments are with 100 runs. The shadows show the 99.9\% confidence intervals. The legends apply to all plots. $\A^{\text{avg}}_{\text{Value}}$ is used.
}\label{fig:german_credit}
\end{minipage}
\end{figure}

\begin{figure}[t]
\begin{minipage}{\linewidth}
\vspace{-2mm}\centering
\includegraphics[width=1.01\linewidth]{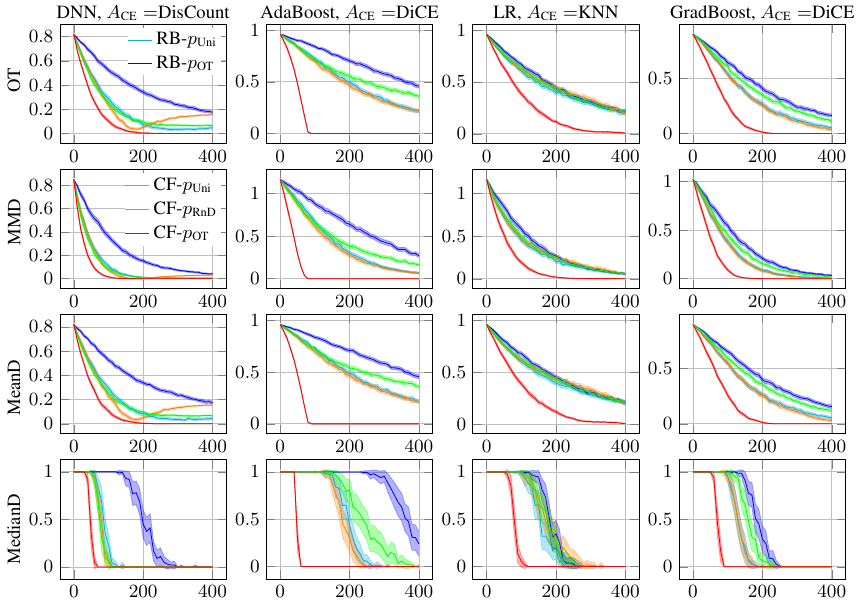}
\caption{[\textit{Hotel Bookings}] $D(f(\vec{z}),\vec{y}^*)$ vs. allowed actions $C$. Experiments are with 100 runs. The shadows show the 99.9\% confidence intervals. The legends apply to all plots. $\A^{\text{max}}_{\text{Value}}$ is used.
}\label{fig:hotel_bookings}
\end{minipage}
\begin{minipage}{\linewidth}
\vspace{3mm}\centering
\includegraphics[width=1.01\linewidth]{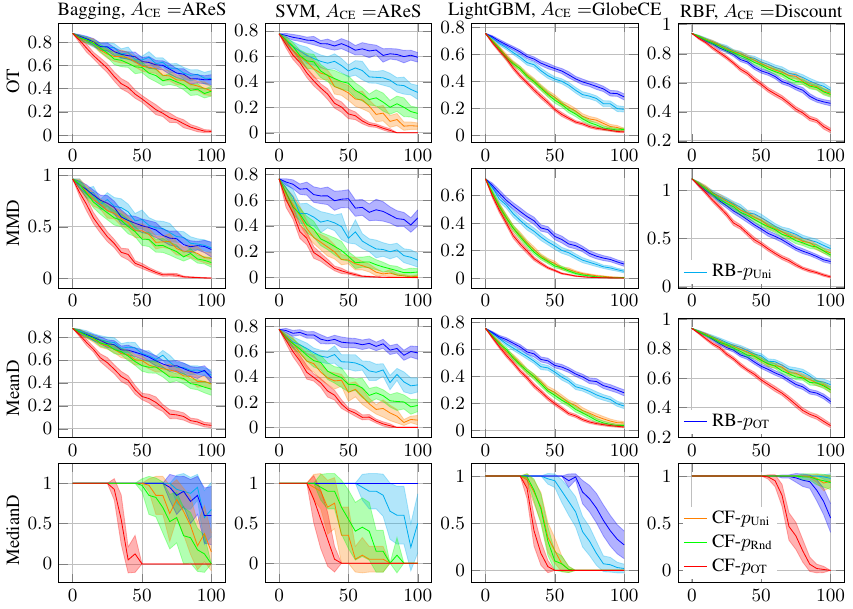}
\caption{[\textit{COMPAS}] $D(f(\vec{z}),\vec{y}^*)$ vs. allowed actions $C$. Experiments are with 100 runs. The shadows show the 99.9\% confidence intervals. The legends apply to all plots. $\A^{\text{max}}_{\text{Value}}$ is used.
}\label{fig:compas}
\end{minipage}
\end{figure}

Observing \figurename{s}~\ref{fig:heloc}--\ref{fig:compas}, CF-$p_\text{Uni}$ generally performs better than RB-$p_\text{Uni}$. Second, consider RB-$p_\text{OT}$ and CF-$p_\text{OT}$ that also differ only in $\A_{\text{Shap}}$, the latter consistently outperforms the former. Hence \textit{RB-SHAP is not suitable for \gls{fa} in \gls{ce}}. 

 We analyze how different Shapley methods affect \gls{fa}, corresponding to lines \ref{alg:cola-A_shap}--\ref{alkg:cola-phi} in \gls{cola}. The shapley methods can be classified into two categories:   First, consider RB-$p_\text{Uni}$ and CF-$p_\text{Uni}$ that differ only in $\A_{\text{Shap}}$. Observing \figurename{s}~\ref{fig:heloc}--\ref{fig:compas}, CF-$p_\text{Uni}$ generally performs better than RB-$p_\text{Uni}$. Second, consider RB-$p_\text{OT}$ and CF-$p_\text{OT}$ that also differ only in $\A_{\text{Shap}}$, the latter consistently outperforms the former. Hence \textit{RB-SHAP is not suitable for \gls{fa} in \gls{ce}}.

 Besides \gls{fa}, the other equally important step of \gls{cola} is line \ref{alg:cola-A_value}, i.e. using the joint probability $p(\vec{x},\vec{r})$ to compose the matrix $\vec{q}$, telling the factual $x$ to which direction to change its features so as to move towards the target model outcome. We observe in \figurename{s}~\ref{fig:heloc}--\ref{fig:compas} that CF-$p_{\text{OT}}$ consistently outperforms all other methods throughout all experiments. Note that all the three methods CF-$p_{\text{Uni}}$, CF-$p_{\text{Rnd}}$, and CF-$p_{\text{OT}}$ provide solution's for the joint probability $\vec{p}$ when the exact alignment between factuals and counterfactuals are unknown. Yet, their performance differ significantly. Simply knowing the \gls{ce} (and its marginal distribution) is insufficient.

 \gls{ot} proves to be exceptionally useful when the alignment information between factual and counterfactual instances is missing or inaccurate. Even when the \gls{ce} algorithm explicitly matches each factual instance to a corresponding counterfactual, it is challenging to justify that the known alignment optimizes performance. This is supported by \figurename~\ref{fig:optimality} in Section \ref{sec:numerical}.

Note that $p_{\text{OT}}$ does not need to be the true joint distribution of $\vec{x}$ and $\vec{r}$ from a data generation perspective. Instead, it should guide \gls{cola} to treat $\vec{x}$ and $\vec{r}$ together for both \gls{fa} and \gls{ce}. 
Furthermore, the QDA column in \figurename~\ref{fig:heloc} shows stableness of \gls{ot}-based methods, while others diverge significantly from the target.
We emphasize that \gls{cola}, however, is \textit{not limited to using \gls{ot}} as $\A_{\text{Prob}}$. As indicated by \figurename~\ref{fig:optimality}, any known best $\vec{p}$ still has non-negligible gap to the global optimality. Devising a better $\A_{\text{Prob}}$ algorithm is hence an interesting topic worth exploration.

\section{Experiments Reproducibility}
\label{sec:reproducibility}
The experiments are conducted on a high performance computing (HPC) cluster, running with four nodes (for the four datasets) in parallel, with each node equipped with two Intel Xeon Processor 2660v3 (10 core, 2.60GHz) and 128 GB memory. The experiment runs approximately 5-10 hours in each node, dependent on the size of the dataset. It is also possible to reproduce the experiment on a laptop, while it costs more computational time generally than using an HPC cluster. 

For the four datasets, the numerical features are standardized, and the categorical features follow either label-encoding or one-hot encoding. Practically, we did not observe remarkable difference between the two encoding methods in terms of \gls{cola}'s performance. The train-test split follows $7:3$.

The optimality baseline as shown in \figurename~\ref{fig:optimality} is solved by Gurobi 11.0.2 \citep{gurobi}. In order to reproduce the optimality baseline, a license of Gurobi is required. Otherwise, we can resort to open-source operations research libraries such as Google-OR tools \citep{google-or}. We remark that solving the \gls{milp} in Appendix~\ref{sec:ect_optimality} is computationally expensive, such that it may only apply to small scale datasets such as German Credit. If one wants to compute the optimality baseline for other datasets, then the number of used features needs to be reduced. 

The hyperparameters of the models used in the experiment are specified as follows. The models Bagging, GP, RBF, RndForest, AdaBoost, GradBoost, and QDA are scikit-learn models \citep{sklearn}, where all hyper-parameters are kept default. The models DNN, SVM, RBF, and LR are implemented by PyTorch \citep{paszke2019pytorch}. The DNN has three layers. The SVM uses the linear kernel. The models XGBoost \citep{chen2016xgboost} and LightGBM \citep{ke2017lightgbm} are used by their scikit-learn interface, with all hyper-parameters kept default.


\section*{The Use of Large Language Models (LLMs)}
We used large language models (LLMs) only as general-purpose assist tools.
Their role was limited to grammar, wording, and light copy-editing of author-written text.
LLMs did not contribute research ideation, modeling choices, experimental design, or results.
All algorithms, proofs, datasets, and analyses were created and verified by the authors.
Any LLM-suggested phrasing was reviewed and edited before inclusion in the paper.
The authors take full responsibility for all content; LLMs are not authors or contributors.
This disclosure complies with the ICLR policy on LLM usage.


\newpage

%

%
\end{document}